\documentclass[11pt]{article}
\usepackage{times,latexsym}
\usepackage{url}
\usepackage[T1]{fontenc}
\usepackage[preprint]{acl}

\usepackage{amsmath,amsfonts,bm}

\def\eqref#1{equation~\ref{#1}}

\def\1{\bm{1}}

\def\ra{{\textnormal{a}}}
\def\rb{{\textnormal{b}}}
\def\rc{{\textnormal{c}}}
\def\rd{{\textnormal{d}}}
\def\re{{\textnormal{e}}}
\def\rf{{\textnormal{f}}}
\def\rg{{\textnormal{g}}}
\def\rh{{\textnormal{h}}}
\def\ri{{\textnormal{i}}}

\DeclareMathAlphabet{\mathsfit}{\encodingdefault}{\sfdefault}{m}{sl}
\SetMathAlphabet{\mathsfit}{bold}{\encodingdefault}{\sfdefault}{bx}{n}

\usepackage{amsmath,amssymb}
\usepackage{booktabs}
\usepackage{multirow}
\usepackage{xcolor}
\usepackage{graphicx}
\usepackage{tikz}
\usetikzlibrary{arrows.meta,positioning,calc,backgrounds,fit}
\newcommand{\rmse}{RMSE}
\newsavebox{\fitbox}
\newcommand{\fitwidth}[1]{\sbox{\fitbox}{#1}\ifdim\wd\fitbox>\linewidth\resizebox{\linewidth}{!}{\usebox{\fitbox}}\else\usebox{\fitbox}\fi}
\newcommand{\nProbeDecision}{74}
\newcommand{\nProbeGDP}{100}

\newcommand{\cwgTextFixedGDP}{\ensuremath{+0.9}}
\newcommand{\cwpTextFixedGDP}{0.064}

\newcommand{\cwgTextGatedGDP}{\ensuremath{-2.6}}

\newcommand{\cwgTextQwenFourteenGDP}{\ensuremath{+0.9}}
\newcommand{\cwpTextQwenFourteenGDP}{0.047}

\newcommand{\guardgGatedGDP}{\ensuremath{-2.3}}

\newcommand{\nDocs}{429}
\newcommand{\nStatements}{221}
\newcommand{\nMinutes}{208}
\newcommand{\nOneSectionMinutes}{6}
\newcommand{\nOrigins}{150}
\newcommand{\nQuarters}{50}

\newcommand{\xcN}{6,688}
\newcommand{\xcMaxDiff}{8\times10^{-5}}
\newcommand{\xcSignAgree}{100}
\newcommand{\robPosOfGDP}{four of six}

\newcommand{\robSigGDP}{none}

\newcommand{\robMinCPI}{\ensuremath{-6.4}}
\newcommand{\robMaxCPI}{\ensuremath{+6.5}}
\newcommand{\robSigCPI}{one}

\newcommand{\pgGated}{0.48}

\newcommand{\pgFixedA}{0.64}

\newcommand{\pgtwFixedA}{0.60}
\newcommand{\pdFixedA}{0.65}

\newcommand{\pgyFixedB}{0.52}

\newcommand{\pgyrawFixedB}{0.70}

\newcommand{\pgyFixedC}{0.61}

\newcommand{\pgyrawFixedC}{0.73}

\newcommand{\pgyFixed}{0.53}

\newcommand{\pgyrawFixed}{0.72}

\newcommand{\pdDatedGated}{0.79}

\newcommand{\pdnDatedGated}{31}

\newcommand{\pgdTQwenOne}{\ensuremath{+0.02}}
\newcommand{\pgdTloQwenOne}{\ensuremath{-0.14}}
\newcommand{\pgdThiQwenOne}{\ensuremath{+0.16}}

\newcommand{\pgPersistence}{0.45}

\newcommand{\pdPersistence}{0.54}

\newcommand{\nModAll}{seven}

\newcommand{\pgqModMin}{0.70}
\newcommand{\pgqModMax}{0.82}
\newcommand{\pgqrModMax}{0.040}
\newcommand{\nYears}{25}

\newcommand{\nModVsGatedBonf}{six}
\newcommand{\nModVsT}{four}
\newcommand{\nModVsTBonf}{none}
\newcommand{\pdunseenFixedA}{0.65}
\newcommand{\pdnunseenFixedA}{31}

\newcommand{\pgdTModMin}{\ensuremath{+0.02}}
\newcommand{\pgdTModMax}{\ensuremath{+0.19}}
\newcommand{\pgyModMin}{0.53}
\newcommand{\pgyModMax}{0.62}
\newcommand{\pgypModMin}{0.075}

\newcommand{\pgLadMax}{0.64}

\newcommand{\nLadBoth}{five}
\newcommand{\nLadSeenHigher}{two}
\newcommand{\nLadSuZero}{five}
\newcommand{\psuMin}{\ensuremath{-0.19}}
\newcommand{\psuMax}{\ensuremath{+0.12}}
\newcommand{\pdModMin}{0.64}
\newcommand{\pdModMax}{0.95}
\newcommand{\nContrasts}{five}
\newcommand{\nContrastPos}{two}
\newcommand{\conNMin}{16}
\newcommand{\conNMax}{20}
\newcommand{\conWide}{1.08}
\newcommand{\conNarrow}{0.61}

\newcommand{\prxFixedAGDP}{\ensuremath{-4.1}}

\newcommand{\prQwenSevenCPI}{\ensuremath{-4.2}}
\newcommand{\prloQwenSevenCPI}{\ensuremath{-6.9}}
\newcommand{\prhiQwenSevenCPI}{\ensuremath{-1.6}}
\newcommand{\prciQwenSevenCPI}{0.022}
\newcommand{\prhlnQwenSevenCPI}{0.006}

\newcommand{\prMistralSevenCPI}{\ensuremath{-3.8}}

\newcommand{\prciMistralSevenCPI}{0.074}
\newcommand{\prhlnMistralSevenCPI}{0.013}

\newcommand{\premAllMinGDP}{\ensuremath{-0.9}}
\newcommand{\premAllMaxGDP}{\ensuremath{+3.4}}
\newcommand{\premXMinGDP}{\ensuremath{-5.1}}
\newcommand{\premXMaxGDP}{\ensuremath{-0.3}}
\newcommand{\premXXMinGDP}{\ensuremath{0.0}}
\newcommand{\premXXMaxGDP}{\ensuremath{+7.7}}
\newcommand{\premRawMinGDP}{\ensuremath{-3.0}}
\newcommand{\premRawMaxGDP}{\ensuremath{+1.3}}

\newcommand{\nCiZeroGDP}{eight}
\newcommand{\mdeMinGDP}{8.8}
\newcommand{\mdeMaxGDP}{13.1}
\newcommand{\nPremAllGDP}{eight}

\newcommand{\holmPremMinGDP}{1.00}

\newcommand{\premAllMinCPI}{\ensuremath{-4.2}}
\newcommand{\premAllMaxCPI}{\ensuremath{+0.9}}

\newcommand{\holmPremMinCPI}{0.24}

\newcommand{\holmCwMin}{1.00}
\newcommand{\nHolmCw}{26}
\newcommand{\mbAllMeanGDP}{1.890}
\newcommand{\mbAllBridgeGDP}{1.913}
\newcommand{\mbAllPGDP}{0.81}

\newcommand{\mbExMeanGDP}{1.273}
\newcommand{\mbExBridgeGDP}{1.428}
\newcommand{\mbExPGDP}{0.19}
\newcommand{\mbExDfmGDP}{1.479}

\newcommand{\mbAllMeanCPI}{2.406}
\newcommand{\mbAllBridgeCPI}{1.148}
\newcommand{\mbAllPCPI}{0.11}

\newcommand{\plpctGated}{22}

\newcommand{\plpctFixed}{92}

\newcommand{\nRotations}{64}

\newcommand{\plpremShare}{8}

\newcommand{\encMinHolm}{0.99}
\newcommand{\encN}{48}

\newcommand{\memLpMin}{0.16}
\newcommand{\memLpMax}{0.49}

\newcommand{\igLast}{November 2022}
\newcommand{\igCounted}{416}
\newcommand{\igLastDay}{2 November 2022}

\newcommand{\igWytMkOlmoSeven}{0.54}
\newcommand{\igWytPOlmoSeven}{0.215}
\newcommand{\igTypeOlmoSeven}{0.83}
\newcommand{\igFoundSt}{183/189}
\newcommand{\igFoundMn}{69/183}

\newcommand{\igPrePct}{68}
\newcommand{\igPostPct}{0}

\title{Reading or Remembering?\\ Measuring Temporal Contamination in Language Models}
\author{Arunkumar V \\ UCE BIT Campus, Anna University \\ Tiruchirappalli, India \\ \texttt{arunkumarv1530@gmail.com} \And Gangadharan G. R. \\ NIT Tiruchirappalli \\ Tiruchirappalli, India \And S. Senthilkumar \\ UCE BIT Campus, Anna University \\ Tiruchirappalli, India}

\begin{document}
\maketitle

\begin{abstract}
Language models are now used to read central-bank text for economic forecasting, but
most were trained on text written after the documents they read. A model reading a 2021
FOMC statement may already know that inflation rose in 2022, so part of its apparent
skill can be memory of the outcome, a temporal form of test-set contamination. We
measure this memory on FOMC statements and minutes from 2000 to 2026 by comparing
modern open models with chronologically consistent models, which are trained only on
text from before a fixed date. Asked about past GDP outcomes given only a date, models
that cannot know the answer still reach an AUC of up to \pgLadMax{} from priors on dates
and eras, so chance is the wrong baseline. Against these memory-free models on the same
questions, most modern models recall which years had strong growth.
Membership-inference scores sort documents by era and document type rather than by
whether a model saw them, as exact counts in an open pretraining corpus show. In
real-time nowcasts, every lookahead premium we estimate on US GDP is smaller than the
forecast comparison can detect at macroeconomic sample sizes, and none on GDP or
inflation survives correction for multiple testing. Lookahead audits should therefore
measure memory directly against memory-free models on the same items.
\end{abstract}

\section{Introduction}\label{sec:intro}

Language models are now used to read central-bank communication for forecasters, and
most of the models in use were trained on text written after the period they are asked
to read. A model that has seen the commentary of 2021 and 2022 does not need to infer
from a 2021 FOMC statement that inflation is about to rise; it can remember that it did.
In machine-learning terms this is temporal contamination, a form of test-set
contamination that runs through time: the outcomes a model is evaluated on appear in
text written after its inputs. Lookahead of
this kind has been documented in financial and economic prediction
\citep{sarkar2024lookahead,glasserman2024lookahead,lopezlira2025memorization,gao2025lookahead,eliseev2026fakedate},
and contamination by release date has been found in benchmark results
\citep{roberts2024cutoff}. Chronologically consistent models, trained only on text
available before a fixed date, now exist to remove it
\citep{he2025chronologically,yan2026datedgpt}.

We want to measure, for a forecasting task, how much of a model's apparent skill is
memory, and whether that memory reaches the forecast when the forecast is made as it
would be in practice: in real time, on the data available that day, and scored against
the first published estimate. The usual checks are poorly suited to this. Scoring a
model against chance assumes that a model which has not seen the outcome guesses at
random. Membership-inference scores assume that the texts a model saw and the texts it
did not see differ only in that respect, an assumption that fails when the two come
from different periods \citep{duan2024mia}. And a comparison of forecast accuracy with
and without the suspect model needs more data than a macroeconomic series provides.

We build such a measurement for the FOMC's statements and minutes from 2000 to 2026 and
for real-time nowcasts of US GDP and CPI growth from 2012 to 2024, and we score each
part of it against a reference other than chance. Date-only probes give a model a date
and ask about an outcome it could know only from later text. The reference is a set of
readers that cannot know the outcome, scored on the same items, and the inference allows
for outcomes that cluster in time and for priors on calendar words. A ladder of fixed
checkpoints from one chronologically consistent family asks whether a model does better
on outcomes its training could have contained than on outcomes it could not. Membership
scores ask whether a model saw a document; we check them against checkpoints with known
cutoffs and against exact counts in one model's indexed training corpus. Finally, a
real-time nowcaster measures what the readings are worth, with a reader behind a
temporal gate as the control, two hard-data baselines, the professional nowcasts, and a
power analysis.

The first three of our six findings are positive results. The other three mark what
these data cannot resolve, and we report them with intervals and, for the forecast, the
smallest effect the test could detect.
\begin{enumerate}
\item Readers that cannot know the outcome do not score 0.5. On the GDP probe they reach
AUCs up to about \pgLadMax{} from priors on dates and eras, so every claim of memory
below compares a model with these readers on the same items.
\item The modern models remember outcomes that the memory-free readers cannot know.
Against the gated reader, \nModVsGatedBonf{} of the \nModAll{} score higher after
adjusting for multiple comparisons. A date retrieves which years grew strongly.
\item Membership scores track era and document type rather than membership. Exact span
counts in OLMo-2's indexed corpus date the end of its FOMC text to \igLastDay{}, a year
before its model card's cutoff, and within a year and a document type Min-K\% does not
separate the documents found from those not found.
\item The probes cannot resolve finer distinctions. Within a year there are too few mixed
outcomes to tell whether a model remembers individual quarters. The modern models'
margins over ChronoGPT-2024, a small chronologically consistent model whose training also
runs past every outcome, do not survive adjustment for multiple comparisons, so whether
the larger models remember more than the small one stays open. The checkpoint ladder's
intervals all include zero and are wider than that gap, which leaves open whether the
small models remember at all.
\item The forecast comparison lacks the power to detect a lookahead premium. The leaky
readers' GDP premiums over the gated reader are a few percent of \rmse{}, below the
\mdeMinGDP{} to \mdeMaxGDP{} percent the test detects with 80\% power, and their sign
depends on whether 2020 or 2021 is left out. They are also measured on a GDP nowcaster
that does not beat a running mean of past growth.
\item Text adds no skill that survives correction for multiple testing, either to the
real-time nowcaster or to the SPF, GDPNow and New York Fed nowcasts.
\end{enumerate}

A lookahead audit of a forecasting system should therefore rest neither on forecast
accuracy, which at macroeconomic sample sizes cannot resolve premiums of a few percent,
nor on membership scores compared across eras, which confound membership with time, but
on date-only probes scored against memory-free readers on the same items, and on exact
corpus counts where a corpus is indexed. We release the corpus, every score and every
script.

\section{Related work}\label{sec:related}

\paragraph{Memorisation and contamination detection}
Memorisation grows log-linearly with model capacity, duplication and prompt length
\citep{carlini2023quantifying}. Detection methods ask whether a text was in the training
corpus: Min-K\% Prob averages the log-probabilities of a text's least likely tokens
\citep{shi2024detecting}, Min-K\%++ normalises them by the next-token distribution
\citep{zhang2025minkpp}, and guided prompts flag contamination when a completion matches a
reference instance \citep{golchin2024timetravel}. Contamination must be measured benchmark
by benchmark \citep{sainz2023nlp}, and answers to Stack Overflow questions are often
collages of memorised text whether or not the question predates training
\citep{caspari2025studying}; \citet{xu2024contamination} and \citet{ravaut2025survey} survey
the methods. \citet{duan2024mia} find that membership inference barely beats chance on
Pile-trained models and trace reported successes to members and non-members drawn from
different periods. Our probes face the same confound, because answer priors track eras. We
ask whether a model knows an outcome that postdates a document, and score each leaky reader
against a gated reader and a persistence rule on the same items rather than against 0.5.

\paragraph{Temporal generalisation and knowledge cutoffs}
Perplexity on text from after the training period worsens with distance from it
\citep{lazaridou2021mind}; conditioning on timestamps improves recall of time-dependent
facts \citep{dhingra2022timeaware}; GPT pass rates on Codeforces and Project Euler problems
reveal contamination by release date \citep{roberts2024cutoff}; and effective cutoffs often
differ from reported ones \citep{cheng2024dated}. We instead use known cutoffs as the
treatment: within one chronologically consistent family, a ladder of fixed checkpoints
shares architecture, size, tokenizer and recipe, and each checkpoint's cutoff divides the
outcomes it could have seen from those it could not.

\paragraph{Chronologically consistent language models}
StoriesLM continues each checkpoint on the following year's news from 1900
\citep{sarkar2024storieslm}, and Time Machine GPT releases point-in-time models
\citep{drinkall2024timemachine}. ChronoBERT and ChronoGPT reach Sharpe ratios in next-day return
prediction comparable to a much larger Llama model \citep{he2025chronologically}, with instruction-tuned versions
following \citep{he2025instruct}. DatedGPT trains twelve 1.3B-parameter models with annual
cutoffs from 2013 to 2024 and finds a return-prediction premium of 26.4 basis points per
standard deviation for models whose training covers the outcome period
\citep{yan2026datedgpt}. These papers evaluate on asset returns. We use ChronoGPT-Instruct as
a gate inside a macroeconomic nowcaster, use fixed checkpoints of ChronoGPT and DatedGPT as
ungated controls, and measure the premium on \nOrigins{} real-time origins over
\nQuarters{} quarters together with the smallest premium the test can detect.

\paragraph{Lookahead bias in economic and financial uses of language models}
Lookahead appears when models predict risk factors from earnings calls and election winners
from biographies \citep{sarkar2024lookahead}; anonymising headlines improves GPT sentiment
trading in-sample, so distraction can outweigh lookahead \citep{glasserman2024lookahead};
models recall economic series from before their cutoff even when told to respect a
historical boundary \citep{lopezlira2025memorization}; and the probability of a non-unknown
answer given only a firm, ticker and date collapses after the cutoff
\citep{gao2025lookahead}. Telling a model to ignore what it knows leaves a 52\% gap to true
ignorance \citep{li2026simulated}, no modern model passes the first fake-date test of
\citet{eliseev2026fakedate} for macroeconomic forecasts, and the PaLM inflation forecasts of
\citet{fariaecastro2023ai} are in-sample for 2019--23. MACROCAST avoids leakage by
pretraining on synthetic series \citep{carriero2026macrocast}, and \citet{ludwig2026llm}
state the general no-leakage condition. Most of this evidence concerns firm-level outcomes or
a model forecasting a series directly. We study a model reading central-bank text as one
input to a real-time nowcaster, price lookahead as a change in RMSE against first release,
and find that the memory a probe detects does not appear as resolvable forecast skill.

\paragraph{Text in macroeconomic nowcasting}
News text adds information that hard indicators miss
\citep{thorsrud2020words,kalamara2022making,ellingsen2022news}. Central-bank text has been
read with topic models \citep{hansen2016shocking,hansen2018transparency}, with machine
learning on staff documents \citep{aruoba2024identifying}, and with encoders trained further
on central-bank text \citep{gambacorta2026cblms}. Our bridge-factor nowcaster follows the
real-time nowcasting literature
\citep{giannone2008nowcasting,banbura2014maximum,higgins2014gdpnow}, in which data vintages
change results \citep{koenig2003realtime,croushore2011frontiers} and model nowcasts beat the
SPF on headline inflation \citep{knotek2017nowcasting}. We drop the two pandemic quarters,
which is acceptable for estimation though it understates forecast uncertainty
\citep{lenza2022var}. These text measures come from topic models, dictionaries or models
trained on a chosen corpus; a general-purpose language model brings a corpus that extends
past the forecast period, so we make the reader's training cutoff the variable under test.

\section{Setup}\label{sec:setup}

\begin{figure*}[t]
\centering
\resizebox{\textwidth}{!}{%
\begin{tikzpicture}[
  font=\small,
  panel/.style={draw=#1!45!black,fill=#1!10,rounded corners=8pt,line width=0.55pt},
  ptitle/.style={anchor=north west,font=\small,inner sep=0pt},
  box/.style={draw=black!85,fill=white,align=center,inner sep=4.5pt,line width=0.5pt,
              font=\footnotesize,minimum height=1.0cm},
  obox/.style={box,fill=black!5,line width=0.75pt},
  note/.style={box,minimum height=0.7cm,draw=black!55,fill=white},
  ar/.style={-{Latex[length=2.1mm,width=1.6mm]},line width=0.65pt},
  dar/.style={ar,dash pattern=on 3.2pt off 2.2pt},
  lab/.style={font=\scriptsize,fill=white,inner sep=1.2pt},
]
\draw[panel=green]  (0,6.9)    rectangle (17.4,10.75);
\node[ptitle,text=green!30!black]  at (0.3,10.55) {Reading the text, one document at a time};
\draw[panel=orange] (0,1.2)    rectangle (7.7,6.15);
\node[ptitle,text=orange!40!black] at (0.3,5.95)  {Measuring memory};
\draw[panel=blue]   (8.5,1.2)  rectangle (17.4,6.15);
\node[ptitle,text=blue!45!black]   at (8.8,5.95)  {Real-time nowcaster at origin $\tau$};
\draw[panel=yellow] (0,-1.35)  rectangle (17.4,0.9);
\node[ptitle,text=yellow!25!black] at (0.3,0.7)   {Evaluation against first release, 2012--2024};
\node[box,minimum width=2.3cm] (docs)  at (1.5,9.55)   {FOMC statements\\and minutes,\\\nDocs{} excerpts};
\node[box,minimum width=2.6cm] (gate)  at (4.95,9.55)  {\textbf{Temporal gate}\\year $y$ is read by\\cutoff 31 Dec $y{-}1$};
\node[box,minimum width=2.6cm] (read)  at (8.6,9.55)   {Reader\\label scores\\$\delta=\log p_{+}-\log p_{-}$};
\node[box,minimum width=2.6cm] (ogate) at (12.3,9.55)  {\textbf{Output gate}\\calibrate on ``N/A'',\\3 paraphrases agree};
\node[obox,minimum width=2.1cm] (z)    at (15.85,9.55) {Policy signal\\$z_\tau\in[-1,1]^5$};
\node[note,align=left] (roster) at (6.75,7.75)
  {Readers: gated ChronoGPT and its fixed checkpoints;\\
   Qwen2.5 1.5B to 14B; OLMo-2 7B, 13B; Mistral-7B;\\
   a dictionary. DatedGPT answers the probes only.};
\node[note] (abst) at (12.3,7.75) {sign disagreement:\\abstain, $z^{(k)}=0$,\\logged};
\draw[ar] (docs) -- (gate);
\draw[ar] (gate) -- (read);
\draw[ar] (read) -- (ogate);
\draw[ar] (ogate) -- (z);
\draw[dar] (ogate.south) -- (abst.north);
\draw[line width=0.5pt,black!55] (read.south) -- (read.south |- roster.north);
\node[box,minimum width=2.9cm] (prompt) at (1.95,4.55) {Date-only prompt\\\nProbeDecision{} rate decisions,\\\nProbeGDP{} GDP quarters};
\node[box,minimum width=2.9cm] (same)   at (5.75,4.55) {Same reader and\\readout, with the\\document left out};
\node[obox,minimum width=6.6cm] (auc)   at (3.85,3.05)
  {AUC with era-robust inference: same quarter word,\\same year, seen against unseen outcomes};
\node[box,minimum width=6.6cm,minimum height=0.8cm] (mem) at (3.85,1.95)
  {Membership scores (Min-K\%, Min-K\%++)\\and exact counts in OLMo-2's corpus};
\draw[ar] (prompt) -- (same);
\draw[ar] (same.south) -- (same.south |- auc.north);
\draw[dar] (roster.south -| same.north) -- node[lab,left=2pt,pos=0.3]{same checkpoints} (same.north);
\node[box,minimum width=2.6cm] (vint)   at (10.35,4.55) {RTDSM vintage\\at $\tau$, 11 series};
\node[box,minimum width=2.3cm] (fill)   at (13.2,4.55)  {Ragged-edge\\AR(2) fill};
\node[box,minimum width=2.1cm] (pca)    at (15.95,4.55) {Block PCA,\\3 factors};
\node[box,minimum width=2.6cm] (struct) at (10.35,3.05) {Text block\\5 signal dimensions,\\standardised};
\node[obox,minimum width=2.1cm] (ridge) at (15.95,3.05) {Ridge bridge\\$\hat y_\tau$};
\node[box,minimum width=6.2cm,minimum height=0.8cm,draw=black!55] (dfm) at (12.95,1.95) {Benchmarks: dynamic factor model, no-text bridge};
\draw[ar] (vint) -- (fill);
\draw[ar] (fill) -- (pca);
\draw[ar] (pca) -- node[lab,right=2pt]{$F_\tau$} (ridge);
\draw[ar] (struct) -- node[lab,above=1.5pt]{$z_\tau$} (ridge);
\draw[dar] (z.south) -- (15.85,6.55) -- node[lab,above=1.5pt,pos=0.5]{$z_\tau$} (8.1,6.55) |- (struct.west);
\node[box,minimum width=3.1cm] (pro)   at (2.2,-0.5)  {Professional nowcasts,\\public values only:\\SPF, GDPNow, NY Fed};
\node[box,minimum width=3.6cm] (loss)  at (8.65,-0.5) {Squared error against\\first-release GDP and CPI,\\\nOrigins{} origins, \nQuarters{} quarters};
\node[obox,minimum width=4.3cm] (tests) at (14.55,-0.5) {premium: leaky against gated,\\quarter-block intervals, Holm,\\power, rotation placebo};
\draw[ar] (pro) -- node[lab,above=1.5pt]{matched by date} (loss);
\draw[ar] (loss) -- (tests);
\draw[ar] (ridge.east) -- (17.1,3.05) -- (17.1,1.05) -- node[lab,pos=0.3,fill=none,above=0.5pt]{$\hat y_\tau$} (8.65,1.05) -- (loss.north);
\end{tikzpicture}%
}
\caption{The pipeline at one forecast origin $\tau$. A temporal gate routes each
FOMC document to the ChronoGPT checkpoint whose training ended the year before it;
ungated readers see every document. An output gate turns next-token label scores
into a five-dimensional policy signal and abstains when three wordings of a
question disagree in sign. The same readers answer date-only memory probes. At each
origin the nowcaster takes one vintage of the Philadelphia Fed Real-Time Data Set
(RTDSM), the eleven indicators as published by that date. It fills the unpublished
months, the ragged edge, with a second-order autoregression on each indicator's monthly
changes (AR(2) fill) and reduces the target's block of indicators to three principal
components (block PCA). A ridge regression of the target on these factors and any text
signal, the ridge bridge, gives the nowcast $\hat y_\tau$. Every nowcast is scored against the first release, the first published
estimate of the quarter.}
\label{fig:architecture}
\end{figure*}
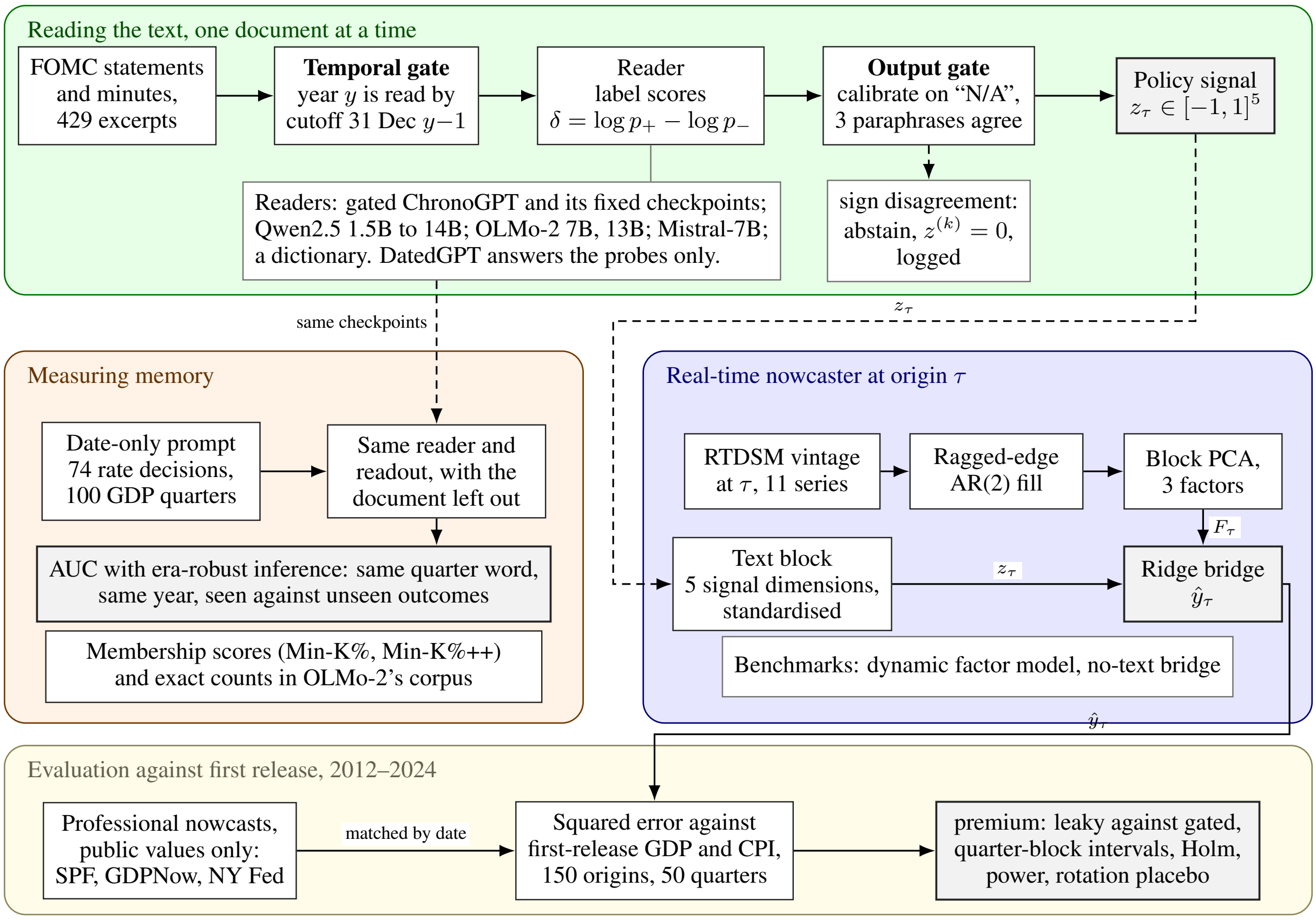

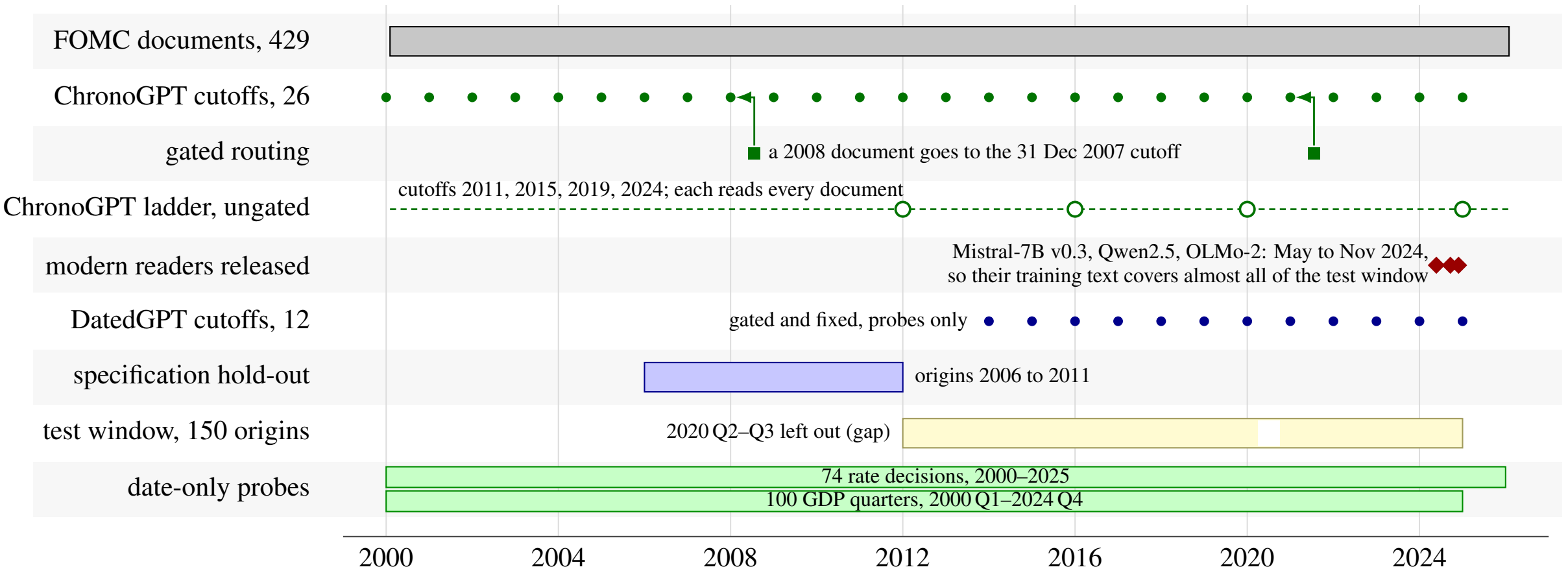
\begin{figure*}[t]
\centering
\resizebox{\textwidth}{!}{%
\begin{tikzpicture}[
  font=\small,
  rowlab/.style={anchor=east,font=\footnotesize,align=right},
  bar/.style={draw=#1!55!black,fill=#1!22,line width=0.45pt},
  ar/.style={-{Latex[length=1.8mm,width=1.4mm]},line width=0.6pt},
  ann/.style={font=\scriptsize,inner sep=1pt},
]
\def\X#1{(3.6+(#1-1999)*0.5)}
\pgfmathsetmacro{\H}{0.17}
\def\ra{6.2}\def\rb{5.55}\def\rc{4.9}\def\rd{4.25}\def\re{3.6}\def\rf{2.95}\def\rg{2.3}\def\rh{1.65}\def\ri{1.0}
\foreach \y in {\ra,\rc,\re,\rg,\ri} \fill[black!3] (0,\y-0.32) rectangle (17.75,\y+0.32);
\foreach \yr in {2000,2004,...,2024} {
  \draw[black!14,line width=0.4pt] ({3.6+(\yr-1999)*0.5},0.45) -- ({3.6+(\yr-1999)*0.5},6.62);
  \node[font=\footnotesize,below] at ({3.6+(\yr-1999)*0.5},0.45) {\yr};
}
\draw[black!70,line width=0.5pt] (3.6,0.45) -- (17.6,0.45);
\node[rowlab] at (3.35,\ra) {FOMC documents, \nDocs{}};
\draw[bar=black] ({3.6+(2000.09-1999)*0.5},\ra-\H) rectangle ({3.6+(2026.08-1999)*0.5},\ra+\H);
\node[rowlab] at (3.35,\rb) {ChronoGPT cutoffs, 26};
\foreach \yr in {2000,...,2025} \fill[green!45!black] ({3.6+(\yr-1999)*0.5},\rb) circle (1.6pt);
\node[rowlab] at (3.35,\rc) {gated routing};
\foreach \d in {2008,2021} {
  \fill[green!45!black] ({3.6+(\d+0.55-1999)*0.5-0.07},\rc-0.07) rectangle ++(0.14,0.14);
  \draw[ar,green!45!black] ({3.6+(\d+0.55-1999)*0.5},\rc+0.09) |- ({3.6+(\d-1999)*0.5+0.06},\rb);
}
\node[ann,anchor=west] at ({3.6+(2008.8-1999)*0.5},\rc) {a 2008 document goes to the 31 Dec 2007 cutoff};
\node[rowlab] at (3.35,\rd) {ChronoGPT ladder, ungated};
\draw[green!45!black,line width=0.6pt,dash pattern=on 2pt off 1.5pt] ({3.6+(2000.09-1999)*0.5},\rd) -- ({3.6+(2026.08-1999)*0.5},\rd);
\foreach \c in {2012,2016,2020,2025} \fill[white,draw=green!45!black,line width=0.8pt] ({3.6+(\c-1999)*0.5},\rd) circle (2.4pt);
\node[ann,anchor=south west] at ({3.6+(2000.2-1999)*0.5},\rd+0.06) {cutoffs 2011, 2015, 2019, 2024; each reads every document};
\node[rowlab] at (3.35,\re) {modern readers released};
\foreach \m in {4.7,8.6,10.9} \fill[red!60!black] ({3.6+(2024+\m/12-1999)*0.5},\re) ++(0,0.1) -- ++(0.1,-0.1) -- ++(-0.1,-0.1) -- ++(-0.1,0.1) -- cycle;
\node[ann,anchor=east,align=right] at ({3.6+(2024.3-1999)*0.5},\re) {Mistral-7B v0.3, Qwen2.5, OLMo-2: May to Nov 2024,\\so their training text covers almost all of the test window};
\node[rowlab] at (3.35,\rf) {DatedGPT cutoffs, 12};
\foreach \yr in {2014,...,2025} \fill[blue!55!black] ({3.6+(\yr-1999)*0.5},\rf) circle (1.6pt);
\node[ann,anchor=east] at ({3.6+(2013.6-1999)*0.5},\rf) {gated and fixed, probes only};
\node[rowlab] at (3.35,\rg) {specification hold-out};
\draw[bar=blue] ({3.6+(2006-1999)*0.5},\rg-\H) rectangle ({3.6+(2012-1999)*0.5},\rg+\H);
\node[ann,anchor=west] at ({3.6+(2012.2-1999)*0.5},\rg) {origins 2006 to 2011};
\node[rowlab] at (3.35,\rh) {test window, \nOrigins{} origins};
\draw[bar=yellow] ({3.6+(2012-1999)*0.5},\rh-\H) rectangle ({3.6+(2025-1999)*0.5},\rh+\H);
\fill[white] ({3.6+(2020.25-1999)*0.5},\rh-\H+0.02) rectangle ({3.6+(2020.75-1999)*0.5},\rh+\H-0.02);
\node[ann,anchor=east] at ({3.6+(2011.8-1999)*0.5},\rh) {2020\,Q2--Q3 left out (gap)};
\node[rowlab] at (3.35,\ri) {date-only probes};
\draw[bar=green] ({3.6+(2000-1999)*0.5},\ri+0.02) rectangle node[font=\scriptsize]{\nProbeDecision{} rate decisions, 2000--2025} ({3.6+(2026-1999)*0.5},\ri+0.26);
\draw[bar=green] ({3.6+(2000-1999)*0.5},\ri-0.26) rectangle node[font=\scriptsize]{\nProbeGDP{} GDP quarters, 2000\,Q1--2024\,Q4} ({3.6+(2025-1999)*0.5},\ri-0.02);
\end{tikzpicture}%
}
\caption{The study on one calendar. ChronoGPT checkpoints end on 31 December of
each year from 1999 to 2024; a document is read by the checkpoint that ends the
year before it. The modern readers were released in 2024, so their training text
covers most of the test window. Four fixed ChronoGPT checkpoints, and three DatedGPT
checkpoints on the probes, form the ungated ladder. Specification selection uses only
origins before the test window.}
\label{fig:design}
\end{figure*}

Figure~\ref{fig:architecture} shows the pipeline at one forecast origin and
Figure~\ref{fig:design} places every window and checkpoint on one calendar.

\paragraph{Corpus}
The corpus is every FOMC statement and set of minutes released from February 2000
to January 2026, \nDocs{}~documents (\nStatements{}~statements and
\nMinutes{}~minutes). The document list and the dates come from the Federal
Reserve's own index: a statement is dated by its press release, a set of minutes by
its release (about three weeks after the meeting since 2005, and after the following
meeting before that). Every reader sees the same excerpts, cut by one fixed rule: a statement runs from the release header to the vote (or to the page footer
when there is no recorded vote); minutes contribute the first 2,500 characters of
the discussion of the outlook and the first 2,500 of the discussion of policy.
Excerpts are at most 5,000 characters. Appendix~\ref{app:corpus} documents the data
cleaning, including the removal of one leak.

\paragraph{Real-time data and targets}
Monthly indicators come from the Philadelphia Fed Real-Time Data Set (RTDSM)
\citep{croushore2001realtime}, which keeps every vintage of each series, that is, the
series as it stood on a given date before later revisions. We use eleven series with
monthly vintages, covering
industrial production, capacity utilisation, payroll employment, hours, housing
starts, three real consumption series and three price indices. Each forecast
origin uses exactly one vintage and is dated the last day of the vintage month, by
which time every release in it is public. Targets are first-release
quarter-on-quarter growth at an annual rate, $400\ln(Y_t/Y_{t-1})$, for real GDP
and for the quarterly average of headline CPI; the first release is the first published
estimate of a quarter, before any revision. The test window has \nOrigins{}
monthly origins over \nQuarters{} quarters from 2012 to 2024; 2020\,Q2 and Q3 are
excluded from estimation and scoring.

\paragraph{Readers}
Table~\ref{tab:readers} in Appendix~\ref{app:readers} lists the readers. The gated reader is the
ChronoGPT-Instruct family \citep{he2025chronologically,he2025instruct}, about 1.55
billion parameters, with a checkpoint ending on 31 December of every year from 1999
to 2024; a document dated in year $y$ goes to the checkpoint ending in $y-1$.
Four of these checkpoints are also run ungated on every document: those ending in
2011, 2015, 2019 and 2024. They form a ladder: each has the same architecture and
recipe as the gated reader, and each could have seen outcomes up to its own cutoff
and no later. Seven modern open models are run ungated. A second chronologically
consistent family, DatedGPT \citep{yan2026datedgpt}, about 1.3 billion parameters
with checkpoints ending in each year from 2013 to 2024, answers the probes only, both
gated and as three fixed checkpoints (2016, 2020, 2024); its gated reader has no
checkpoint for the 2000--2011 training window of the nowcaster, and we did not run its
fixed checkpoints on the documents. A dictionary reader is reported for reference;
its phrase lists were written with knowledge of later Fed language, so it is not
memory-free.

\paragraph{Reading a document}
For an excerpt $d$, a dimension $k$ (policy stance, inflation risk, growth risk,
uncertainty, forward guidance) and one of three wordings $j$ of the question, the
reader is asked a question with two one-word answers, such as whether policy is
\emph{more} or \emph{less} restrictive than neutral (Appendix~\ref{app:prompts}).
We read the next-token probabilities of the two answers instead of generating
text. An answer's probability sums its single-token forms, and the raw score is
$\delta_{dkj}=\log p(\text{pos}\mid d,k,j)-\log p(\text{neg}\mid d,k,j)$.

\paragraph{The output gate}
Two corrections precede any use of a score. Each reader's score on a content-free
input, the same prompt with the document replaced by ``N/A'', is subtracted for the
same reader, checkpoint, dimension and wording \citep{zhao2021calibrate}. The
calibrated scores of the three wordings must then agree in sign; if they do, the
signal for dimension $k$ is $\frac13\sum_j(2\sigma(\tilde\delta_{dkj})-1)$, and if
not, the gate abstains, sets it to zero and logs the three scores. The gate is there
to flag readings a model cannot give consistently; Section~\ref{sec:results:value} and
Appendix~\ref{app:gate} report how it changes accuracy.

\section{Measuring memory}\label{sec:memory}

We measure memory three ways, each with its own null, before asking what memory
is worth in a forecast.

\paragraph{Date-only probes}
A reader is given a date and nothing else, a two-answer form of the date-only recall
queries of \citet{gao2025lookahead}. In the decision probe it is told that the
FOMC changed its target on that date and asked whether it raised or lowered it
(\nProbeDecision{} changes, 2000--2025). In the GDP probe it is asked whether
first-release real GDP growth in a given quarter was above or below 2 percent at an
annual rate (\nProbeGDP{} quarters, 2000--2024). A reader that has not seen the
outcome can only guess, but a guess need not score an AUC of 0.5: answer priors
can line up with eras in which one outcome was common, and a gated reader's
checkpoint changes every January. The null is therefore set by two other readers scored
on the same items: the gated reader and a persistence rule that uses only what was
public before the item's year began. Probe items cluster in time, so inference
respects the clustering: $p$-values come from rotating the labels against the
scores in date order (a circular shift, which keeps the dependence in both
series), and intervals for AUC differences from resampling calendar years. We also
compare quarters with the same quarter number across years, tested by rotating the
years of the outcomes against the scores, and quarters within the same year, which
removes most between-era comparisons (a fourth quarter is dated by its release in the
following January, so it is read by the next gated checkpoint).

\paragraph{The checkpoint ladder}
The gated reader mixes 26 checkpoints, and its 2024 sibling has seen more text in
total, so the gated-against-ungated contrast differs in more than lookahead. The
ladder separates the two. Each fixed checkpoint, ending in 2011, 2015, 2019 or
2024, reads every document and answers every probe. For each, an item is
\emph{seen} if its outcome was public by the checkpoint's cutoff and \emph{unseen}
otherwise. Memory predicts that a checkpoint does better on what it could have seen
than on what it could not; general capability, recency and checkpoint mixing do not
predict that split within one checkpoint.

\paragraph{Membership}\label{sec:membership}
Probes ask whether a model knows an outcome; membership scores ask whether it saw a
text. For every excerpt and reader we compute Min-K\% Prob, the mean of the lowest
$k=20$ percent of token log-probabilities of the excerpt read as plain text
\citep{shi2024detecting}, and Min-K\%++, which standardises each token's
log-probability by the model's own next-token distribution
\citep{zhang2025minkpp}. Membership scores are known to fail when members and
non-members come from different periods \citep{duan2024mia}, as ours do. For each fixed ChronoGPT checkpoint, documents released after its cutoff are
certainly not members and documents released before it possibly are; for OLMo-2, whose
pretraining corpus is indexed, we count exact spans of each excerpt in the corpus.

\subsection{Results: what the readers remember}\label{sec:results:memory}

\paragraph{Memory-free readers do not score 0.5}
Table~\ref{tab:probes2} gives the GDP probe for every reader. The readers that cannot
know an outcome are the right null, and they are not at chance: the gated ChronoGPT
reader scores \pgGated{}, the persistence rule \pgPersistence{}, and checkpoints that
cannot have seen most of their items score as high as \pgFixedA{} (ChronoGPT with a
2011 cutoff, which scores \pgtwFixedA{} on the 2012--2024 items it cannot know). AUCs
up to about \pgLadMax{} therefore arise from priors on dates and eras, and a modern
reader's AUC is evidence of memory only where it exceeds these readers on the same
items, with intervals that respect how outcomes cluster in years.

\paragraph{The modern readers exceed them}
Against the gated ChronoGPT reader on the same items, the intervals of \nModVsGatedBonf{}
of the \nModAll{} modern readers exclude zero after adjusting for seven comparisons. The
sharper comparison is with ChronoGPT-2024, a 1.55-billion-parameter reader whose training
also runs past every outcome: the modern readers, of 1.5 to 14 billion parameters, score
\pgdTModMin{} to \pgdTModMax{} higher, the 95\% intervals of \nModVsT{} of the \nModAll{}
exclude zero (Qwen2.5-3B, Qwen2.5-14B, OLMo-2-7B and Mistral-7B), and after adjusting for
seven comparisons \nModVsTBonf{} do. Qwen2.5-1.5B, the one reader of comparable size,
scores \pgdTQwenOne{} $[\pgdTloQwenOne{}, \pgdThiQwenOne{}]$. On pairs of quarters with the
same quarter word across years, which removes any prior on the words ``first'' to
``fourth'', the modern readers score \pgqModMin{} to \pgqModMax{}; rotating the years of
the outcomes against the scores, which keeps how outcomes cluster across years, puts every
one of them at the smallest $p$ the \nYears{} years allow (\pgqrModMax{}), but also puts
ChronoGPT-2015 there. Within a year, after each reader's mean answer for each quarter word
is removed, the modern readers score \pgyModMin{} to \pgyModMax{} (smallest $p$
\pgypModMin{}); few years mix outcomes, so quarter-level memory is not resolvable
(Appendix~\ref{app:probes}).

\paragraph{The ladder cannot resolve memory of this size}
For the \nLadBoth{} checkpoints with both kinds of item, the seen-minus-unseen
difference in AUC runs from \psuMin{} to \psuMax{}, \nLadSeenHigher{}
are positive, and all \nLadSuZero{} intervals include zero (Table~\ref{tab:ladder},
Figure~\ref{fig:ladder} in Appendix~\ref{app:figs}). The seen and unseen items come from different eras, so the
cleaner contrast holds the items fixed and compares consecutive checkpoints on the
outcomes released between their cutoffs: \nContrastPos{} of the \nContrasts{} contrasts
are positive, on \conNMin{} to \conNMax{} items each, with intervals from \conNarrow{}
to \conWide{} wide (Table~\ref{tab:contrasts}). The ladder cannot detect memory of the
size the modern readers show over ChronoGPT-2024, so it neither confirms nor excludes
memory in the small chronologically consistent models.

\paragraph{The decision probe fails its null}
Readers that cannot know the outcome score well above 0.5 on the decision probe: the gated
DatedGPT reader scores \pdDatedGated{} on its \pdnDatedGated{} items, and ChronoGPT with a
2011 cutoff scores \pdunseenFixedA{} on the \pdnunseenFixedA{} rate changes after its
cutoff. We do not know which prior they exploit (the persistence rule, the direction of the
last change before the year, scores only \pdPersistence{}), so the probe cannot separate
memory from such priors. The modern readers score \pdModMin{} to \pdModMax{}. We report
the probe in Appendix~\ref{app:probes} and draw no conclusion from it.

\begin{table*}[t]
\centering
\caption{The GDP probe: was first-release growth in a quarter above 2 percent, given
only the quarter? AUC of the raw label score. Differences against the reader's own gated
family (ChronoGPT or DatedGPT) and against ChronoGPT-2024, on the same items, with 95\%
intervals from resampling years. ``Same quarter word'': pairs of quarters with the same
quarter number; $p$ from rotating the years of the outcomes against the scores, whose
floor is $1/\nYears{}$. DatedGPT's gated reader covers items from 2014 on.}
\label{tab:probes2}
\fitwidth{\scriptsize\begin{tabular}{@{}l r r l l r r@{}}
\toprule
& & & \multicolumn{2}{c}{AUC difference, 95\% year-block interval} & \multicolumn{2}{c}{Same quarter word} \\
\cmidrule(lr){4-5}\cmidrule(lr){6-7}
Reader & $n$ & AUC & against own gated reader & against ChronoGPT-2024 & AUC & rotation $p$ \\
\midrule
ChronoGPT, gated & 100 & 0.48 & n/a & $-0.15$ $[-0.34, +0.02]$ & 0.48 & 0.680 \\
ChronoGPT, cutoff 2011 & 100 & 0.64 & $+0.15$ $[-0.01, +0.32]$ & $+0.00$ $[-0.11, +0.11]$ & 0.64 & 0.120 \\
ChronoGPT, cutoff 2015 & 100 & 0.62 & $+0.14$ $[-0.00, +0.30]$ & $-0.01$ $[-0.17, +0.15]$ & 0.58 & 0.040 \\
ChronoGPT, cutoff 2019 & 100 & 0.51 & $+0.03$ $[-0.15, +0.24]$ & $-0.12$ $[-0.29, +0.04]$ & 0.43 & 0.840 \\
ChronoGPT, cutoff 2024 & 100 & 0.64 & $+0.15$ $[-0.03, +0.34]$ & n/a & 0.56 & 0.200 \\
\addlinespace
DatedGPT, gated & 45 & 0.54 & n/a & $-0.09$ $[-0.46, +0.19]$ & 0.56 & 0.333 \\
DatedGPT, cutoff 2016 & 100 & 0.59 & $+0.12$ $[-0.06, +0.34]$ & $-0.04$ $[-0.17, +0.09]$ & 0.57 & 0.120 \\
DatedGPT, cutoff 2020 & 100 & 0.36 & $-0.23$ $[-0.44, +0.04]$ & $-0.27$ $[-0.48, -0.07]$ & 0.40 & 1.000 \\
DatedGPT, cutoff 2024 & 100 & 0.44 & $-0.06$ $[-0.27, +0.13]$ & $-0.20$ $[-0.41, +0.02]$ & 0.45 & 0.960 \\
\addlinespace
Qwen2.5-1.5B & 100 & 0.66 & $+0.17$ $[+0.00, +0.33]$ & $+0.02$ $[-0.14, +0.16]$ & 0.70 & 0.040 \\
Qwen2.5-3B & 100 & 0.76 & $+0.28$ $[+0.13, +0.41]$ & $+0.12$ $[+0.01, +0.24]$ & 0.77 & 0.040 \\
Qwen2.5-7B & 100 & 0.79 & $+0.31$ $[+0.14, +0.47]$ & $+0.16$ $[-0.02, +0.31]$ & 0.81 & 0.040 \\
Qwen2.5-14B & 100 & 0.79 & $+0.31$ $[+0.15, +0.47]$ & $+0.16$ $[+0.04, +0.28]$ & 0.79 & 0.040 \\
OLMo-2-7B & 100 & 0.78 & $+0.30$ $[+0.15, +0.45]$ & $+0.14$ $[+0.01, +0.27]$ & 0.78 & 0.040 \\
OLMo-2-13B & 100 & 0.75 & $+0.27$ $[+0.11, +0.41]$ & $+0.11$ $[-0.05, +0.26]$ & 0.73 & 0.040 \\
Mistral-7B & 100 & 0.83 & $+0.35$ $[+0.20, +0.50]$ & $+0.19$ $[+0.04, +0.32]$ & 0.82 & 0.040 \\
Persistence rule & 100 & 0.45 & $-0.04$ $[-0.21, +0.14]$ & n/a & 0.44 & 0.760 \\
\bottomrule
\end{tabular}
}
\end{table*}

\begin{table*}[t]
\centering
\caption{The checkpoint ladder. ``Seen'': probe items whose outcome was public by the
checkpoint's cutoff, and nowcast quarters whose first release was; ``unseen'': the rest.
Intervals from resampling years (probes) and quarters (nowcasts). The DatedGPT
checkpoints answered the probes only.}
\label{tab:ladder}
\fitwidth{\scriptsize\begin{tabular}{@{}l ll l ll@{}}
\toprule
& \multicolumn{3}{c}{GDP probe AUC, 95\% year-block interval} & \multicolumn{2}{c}{GDP premium over gated, \%, 95\% interval} \\
\cmidrule(lr){2-4}\cmidrule(lr){5-6}
Checkpoint & seen ($n$) & unseen ($n$) & seen minus unseen & seen (origins) & unseen (origins) \\
\midrule
ChronoGPT, cutoff 2011 & 0.71 $[0.54, 0.86]$ (47) & 0.59 $[0.44, 0.73]$ (53) & $+0.12$ $[-0.13, +0.34]$ & n/a (0) & $+0.6$ $[-6.7, +10.2]$ (150) \\
ChronoGPT, cutoff 2015 & 0.60 $[0.44, 0.78]$ (63) & 0.66 $[0.52, 0.84]$ (37) & $-0.06$ $[-0.33, +0.15]$ & $-6.6$ $[-20.0, +4.1]$ (45) & $+0.1$ $[-6.8, +10.2]$ (105) \\
ChronoGPT, cutoff 2019 & 0.51 $[0.38, 0.66]$ (79) & 0.70 $[0.53, 0.89]$ (21) & $-0.19$ $[-0.46, +0.04]$ & $-1.5$ $[-17.1, +11.1]$ (93) & $+0.2$ $[-6.6, +10.4]$ (57) \\
ChronoGPT, cutoff 2024 & 0.64 $[0.53, 0.76]$ (99) & n/a (1) & n/a & $+3.4$ $[-2.9, +12.3]$ (147) & n/a (3) \\
DatedGPT, cutoff 2016 & 0.54 $[0.39, 0.71]$ (67) & 0.66 $[0.50, 0.86]$ (33) & $-0.12$ $[-0.39, +0.11]$ & not run & not run \\
DatedGPT, cutoff 2020 & 0.39 $[0.24, 0.54]$ (83) & 0.33 $[0.00, 0.70]$ (17) & $+0.06$ $[-0.31, +0.45]$ & not run & not run \\
DatedGPT, cutoff 2024 & 0.44 $[0.32, 0.57]$ (99) & n/a (1) & n/a & not run & not run \\
\bottomrule
\end{tabular}
}
\end{table*}

\paragraph{Membership}
Membership scores cannot be validated on the chronologically consistent checkpoints,
because no one knows which documents released before a cutoff were in the training
data. The comparison does show the era
confound (Table~\ref{tab:membership}): the mean log-probability ranks pre-cutoff
documents as no more familiar than post-cutoff ones, usually less (AUC \memLpMin{} to
\memLpMax{}), because the Fed's language changes over the decades. For OLMo-2 we count up to
three novel 14-word spans of each excerpt in its indexed stage-1 pretraining corpus with
infini-gram \citep{liu2024infinigram}; a span is novel when no earlier FOMC
document contains it, because statements repeat whole sentences. Of the
\igCounted{} documents with a novel span, \igPrePct{}\% of those released up to 2021 are
found and \igPostPct{}\% of those released from 2023; the last one found is the
statement of \igLastDay{}, a year before the model card's cutoff. An absent span may have
been seen in another form, and a statement quoted in the news may be found through the
quotation, so the counts measure exposure to the text rather than to the document.
Statements are almost all found (\igFoundSt{} up to 2022) and minutes much less often
(\igFoundMn{}), and Min-K\% separates the two document types (AUC \igTypeOlmoSeven{} for
OLMo-2-7B). Within a year and a document type, where neither era nor type can help, Min-K\%
does not separate the documents found from those not found (AUC \igWytMkOlmoSeven{},
permutation $p=\igWytPOlmoSeven{}$). On these data, membership scores track era and
document type rather than membership.

\section{Measuring what memory is worth}\label{sec:value}

\paragraph{The real-time nowcaster}
We price memory by putting each reader's signal into a real-time nowcaster, which
estimates growth in the current quarter before the official figure is published. At each origin the nowcaster takes one vintage of the eleven indicators.
Indicators are released with different delays, so the latest months are missing for
some series; this uneven end of the panel is the ragged edge. The nowcaster fills it
with forecasts from an autoregression of order two, AR(2), on each indicator's own
monthly changes, averages to the quarter, and forms growth. It then reduces the
target's own block of indicators (real activity for GDP, prices for CPI) to its first
three principal components, which we call block PCA, and regresses the target on these
three factors with a ridge penalty. This ridge regression is the bridge, the usual name
for a model that links monthly indicators to a quarterly target. It is estimated on the
quarters from 2000\,Q1 whose first release is out by the origin. A text model adds the five dimensions of the
latest FOMC document published on or before the origin; training quarters take the
document available at the same position in their own quarter. Every regressor is
standardised by its training mean and standard deviation, so the penalty treats
every reader alike. The fill and the indicator blocks were chosen among six
specifications on a 2006--2011 hold-out; all six are reported on both windows
(Appendix~\ref{app:specs}), and the order in which they were tried is stated there.

\paragraph{Baselines}
Text is only worth something if the hard-data model is competent, so we report two
further benchmarks. The first is a mixed-frequency dynamic factor model in the style of
\citet{banbura2014maximum}, fit by expectation maximisation on the same vintages
with the ragged edge left as missing data, which this estimator handles directly, and
with factor number and order chosen on the hold-out. The second is the set of professional nowcasts practitioners use: the Survey
of Professional Forecasters, GDPNow \citep{higgins2014gdpnow} and the New York Fed
Staff Nowcast \citep{bok2018nowcasting}, each used only where its value was public
on the date it carries: GDPNow from July 2014, when it went online, and the New
York Fed nowcast during its live publication, April 2016 to September 2021 and from
September 2023. We also ask directly whether a text signal corrects the
professional nowcasts, by regressing their first-release error on the signal out
of sample.

\paragraph{The lookahead premium}
The lookahead premium of a leaky reader is the percentage reduction in nowcast
\rmse{} of its text model against the gated reader's text model, on the same
origins. Three origins share each target quarter. The primary test is the
Coroneo--Iacone test \citep{coroneo2020comparing}, which compares the accuracy of two
forecasts and is built for small samples with autocorrelated errors. We run it two-sided, with
a weighted-periodogram estimate of the long-run variance, bandwidth
$m=\lfloor T^{1/3}\rfloor$ for $T=\nOrigins{}$ origins and a Student-$t$ reference with
$2m$ degrees of freedom. As checks, we compute the Harvey--Leybourne--Newbold test \citep{harvey1997testing},
a small-sample correction of the standard test of equal forecast accuracy, run with a
horizon of three origins, and a 95\% interval from resampling quarters. The gain of a text model over the model without text is a nested
comparison, because the model without text is the text model with its text
coefficients set to zero. We test it with the one-sided Clark--West test
\citep{clark2007approximately}, which adjusts the loss difference for the noise the
larger model adds by estimating coefficients that are zero under the null. Holm's
method, a step-down version of the Bonferroni correction, adjusts each family of tests
for multiplicity: the
premium tests within each target, the \nHolmCw{} text-against-no-text tests across both
targets, and the \encN{} tests of text on top of the professional nowcasts. The minimum detectable effect is the premium the
primary test would detect with 80\% power, from a quarter-block bootstrap standard
error of the mean loss difference and Student-$t$ quantiles with $2m$ degrees of
freedom. We report the premium leaving out each year in turn, and a placebo for the
gated and ChronoGPT-2024 readers: the sequence of document signals is rotated against
the calendar, which keeps the signal's distribution and persistence and breaks its
alignment with outcomes, legitimate information included.

\subsection{Results: what memory is worth}\label{sec:results:value}

\paragraph{A baseline without GDP skill}
We want to know what text adds to a competent nowcaster, and on GDP the hard-data
models are not competent. Over all origins the bridge scores \mbAllBridgeGDP{} and a
running mean of past first releases \mbAllMeanGDP{} ($p=\mbAllPGDP{}$); without
2020--2021 the running mean scores \mbExMeanGDP{} against \mbExBridgeGDP{} for the
bridge ($p=\mbExPGDP{}$) and \mbExDfmGDP{} for the dynamic factor model. Any GDP
premium is therefore measured on a base with no skill over the historical mean. On
CPI the bridge is much more accurate than the running mean, \mbAllBridgeCPI{} against
\mbAllMeanCPI{} ($p=\mbAllPCPI{}$).

\paragraph{Text against no text}
Table~\ref{tab:value} puts every reader's signal into the same nowcaster. Against the
bridge without text, the gated reader's signal changes GDP \rmse{} by
\cwgTextGatedGDP{}\%, the ChronoGPT-2024 signal by \cwgTextFixedGDP{}\% (Clark--West
$p=\cwpTextFixedGDP{}$), and the best modern reader, Qwen2.5-14B, by
\cwgTextQwenFourteenGDP{}\% ($p=\cwpTextQwenFourteenGDP{}$). After Holm adjustment over
the \nHolmCw{} text-against-no-text tests the smallest adjusted $p$ is \holmCwMin{}.
On CPI no reader's signal improves the bridge.

\begin{figure*}[t]
\centering
\fitwidth{\begin{tikzpicture}[font=\scriptsize]
\draw[black!10] (0.000,0) -- (0.000,3.570);
\node[below] at (0.000,0) {-10};
\draw[black!10] (1.120,0) -- (1.120,3.570);
\node[below] at (1.120,0) {-5};
\draw[black!45] (2.240,0) -- (2.240,3.570);
\node[below] at (2.240,0) {0};
\draw[black!10] (3.360,0) -- (3.360,3.570);
\node[below] at (3.360,0) {5};
\draw[black!10] (4.480,0) -- (4.480,3.570);
\node[below] at (4.480,0) {10};
\draw[black!10] (5.600,0) -- (5.600,3.570);
\node[below] at (5.600,0) {15};
\node at (2.800,-0.55) {premium over the gated reader, \%};
\node[font=\footnotesize] at (2.800,3.820) {US GDP};
\node[anchor=east] at (-0.1,0.315) {Mistral-7B};
\draw[black!70,line width=0.6pt] (1.539,0.315) -- (4.415,0.315);
\fill[black] (2.782,0.315) circle (1.8pt);
\draw[red!65!black,line width=0.7pt] (1.978,0.225) -- (1.978,0.405);
\draw[blue!60!black,line width=0.7pt] (3.235,0.225) -- (3.235,0.405);
\node[anchor=east] at (-0.1,0.735) {OLMo-2-13B};
\draw[black!70,line width=0.6pt] (1.485,0.735) -- (4.941,0.735);
\fill[black] (2.925,0.735) circle (1.8pt);
\draw[red!65!black,line width=0.7pt] (1.913,0.645) -- (1.913,0.825);
\draw[blue!60!black,line width=0.7pt] (3.694,0.645) -- (3.694,0.825);
\node[anchor=east] at (-0.1,1.155) {OLMo-2-7B};
\draw[black!70,line width=0.6pt] (0.981,1.155) -- (3.745,1.155);
\fill[black] (2.229,1.155) circle (1.8pt);
\draw[red!65!black,line width=0.7pt] (1.537,1.065) -- (1.537,1.245);
\draw[blue!60!black,line width=0.7pt] (2.632,1.065) -- (2.632,1.245);
\node[anchor=east] at (-0.1,1.575) {Qwen2.5-14B};
\draw[black!70,line width=0.6pt] (1.848,1.575) -- (4.619,1.575);
\fill[black] (3.006,1.575) circle (1.8pt);
\draw[red!65!black,line width=0.7pt] (2.173,1.485) -- (2.173,1.665);
\draw[blue!60!black,line width=0.7pt] (3.660,1.485) -- (3.660,1.665);
\node[anchor=east] at (-0.1,1.995) {Qwen2.5-7B};
\draw[black!70,line width=0.6pt] (1.467,1.995) -- (4.715,1.995);
\fill[black] (2.908,1.995) circle (1.8pt);
\draw[red!65!black,line width=0.7pt] (1.756,1.905) -- (1.756,2.085);
\draw[blue!60!black,line width=0.7pt] (3.429,1.905) -- (3.429,2.085);
\node[anchor=east] at (-0.1,2.415) {Qwen2.5-3B};
\draw[black!70,line width=0.6pt] (0.000,2.415) -- (3.851,2.415);
\fill[black] (2.043,2.415) circle (1.8pt);
\draw[red!65!black,line width=0.7pt] (1.089,2.325) -- (1.089,2.505);
\draw[blue!60!black,line width=0.7pt] (2.244,2.325) -- (2.244,2.505);
\node[anchor=east] at (-0.1,2.835) {Qwen2.5-1.5B};
\draw[black!70,line width=0.6pt] (1.344,2.835) -- (3.945,2.835);
\fill[black] (2.547,2.835) circle (1.8pt);
\draw[red!65!black,line width=0.7pt] (1.711,2.745) -- (1.711,2.925);
\draw[blue!60!black,line width=0.7pt] (2.892,2.745) -- (2.892,2.925);
\node[anchor=east] at (-0.1,3.255) {ChronoGPT-2024, ungated};
\draw[black!70,line width=0.6pt] (1.635,3.255) -- (5.058,3.255);
\fill[black] (3.002,3.255) circle (1.8pt);
\draw[red!65!black,line width=0.7pt] (2.043,3.165) -- (2.043,3.345);
\draw[blue!60!black,line width=0.7pt] (3.974,3.165) -- (3.974,3.345);
\draw[black!10] (6.500,0) -- (6.500,3.570);
\node[below] at (6.500,0) {-10};
\draw[black!10] (7.620,0) -- (7.620,3.570);
\node[below] at (7.620,0) {-5};
\draw[black!45] (8.740,0) -- (8.740,3.570);
\node[below] at (8.740,0) {0};
\draw[black!10] (9.860,0) -- (9.860,3.570);
\node[below] at (9.860,0) {5};
\draw[black!10] (10.980,0) -- (10.980,3.570);
\node[below] at (10.980,0) {10};
\draw[black!10] (12.100,0) -- (12.100,3.570);
\node[below] at (12.100,0) {15};
\node at (9.300,-0.55) {premium over the gated reader, \%};
\node[font=\footnotesize] at (9.300,3.820) {US CPI};
\draw[black!70,line width=0.6pt] (7.333,0.315) -- (8.449,0.315);
\fill[black] (7.889,0.315) circle (1.8pt);
\draw[red!65!black,line width=0.7pt] (8.059,0.225) -- (8.059,0.405);
\draw[blue!60!black,line width=0.7pt] (7.893,0.225) -- (7.893,0.405);
\draw[black!70,line width=0.6pt] (7.983,0.735) -- (8.937,0.735);
\fill[black] (8.512,0.735) circle (1.8pt);
\draw[red!65!black,line width=0.7pt] (8.675,0.645) -- (8.675,0.825);
\draw[blue!60!black,line width=0.7pt] (8.516,0.645) -- (8.516,0.825);
\draw[black!70,line width=0.6pt] (7.582,1.155) -- (9.022,1.155);
\fill[black] (8.299,1.155) circle (1.8pt);
\draw[red!65!black,line width=0.7pt] (8.494,1.065) -- (8.494,1.245);
\draw[blue!60!black,line width=0.7pt] (8.308,1.065) -- (8.308,1.245);
\draw[black!70,line width=0.6pt] (8.102,1.575) -- (9.170,1.575);
\fill[black] (8.677,1.575) circle (1.8pt);
\draw[red!65!black,line width=0.7pt] (8.910,1.485) -- (8.910,1.665);
\draw[blue!60!black,line width=0.7pt] (8.673,1.485) -- (8.673,1.665);
\draw[black!70,line width=0.6pt] (7.201,1.995) -- (8.375,1.995);
\fill[black] (7.806,1.995) circle (1.8pt);
\draw[red!65!black,line width=0.7pt] (8.046,1.905) -- (8.046,2.085);
\draw[blue!60!black,line width=0.7pt] (7.824,1.905) -- (7.824,2.085);
\draw[black!70,line width=0.6pt] (7.920,2.415) -- (9.002,2.415);
\fill[black] (8.518,2.415) circle (1.8pt);
\draw[red!65!black,line width=0.7pt] (8.760,2.325) -- (8.760,2.505);
\draw[blue!60!black,line width=0.7pt] (8.536,2.325) -- (8.536,2.505);
\draw[black!70,line width=0.6pt] (8.321,2.835) -- (9.535,2.835);
\fill[black] (8.951,2.835) circle (1.8pt);
\draw[red!65!black,line width=0.7pt] (9.094,2.745) -- (9.094,2.925);
\draw[blue!60!black,line width=0.7pt] (8.933,2.745) -- (8.933,2.925);
\draw[black!70,line width=0.6pt] (8.182,3.255) -- (9.119,3.255);
\fill[black] (8.711,3.255) circle (1.8pt);
\draw[red!65!black,line width=0.7pt] (8.648,3.165) -- (8.648,3.345);
\draw[blue!60!black,line width=0.7pt] (8.718,3.165) -- (8.718,3.345);
\fill[black] (0.1,4.320) circle (1.8pt) node[right=2pt] {all quarters, with 95\% quarter-block interval};
\draw[red!65!black,line width=0.7pt] (6.3,4.230) -- ++(0,0.18) node[right=2pt,black,pos=0.5] {without 2021};
\draw[blue!60!black,line width=0.7pt] (8.6,4.230) -- ++(0,0.18) node[right=2pt,black,pos=0.5] {without 2020};
\end{tikzpicture}}
\caption{Lookahead premium of each leaky reader over the gated reader, all quarters
(dot, 95\% quarter-block interval) and without 2021 (bar).}
\label{fig:forest}
\end{figure*}
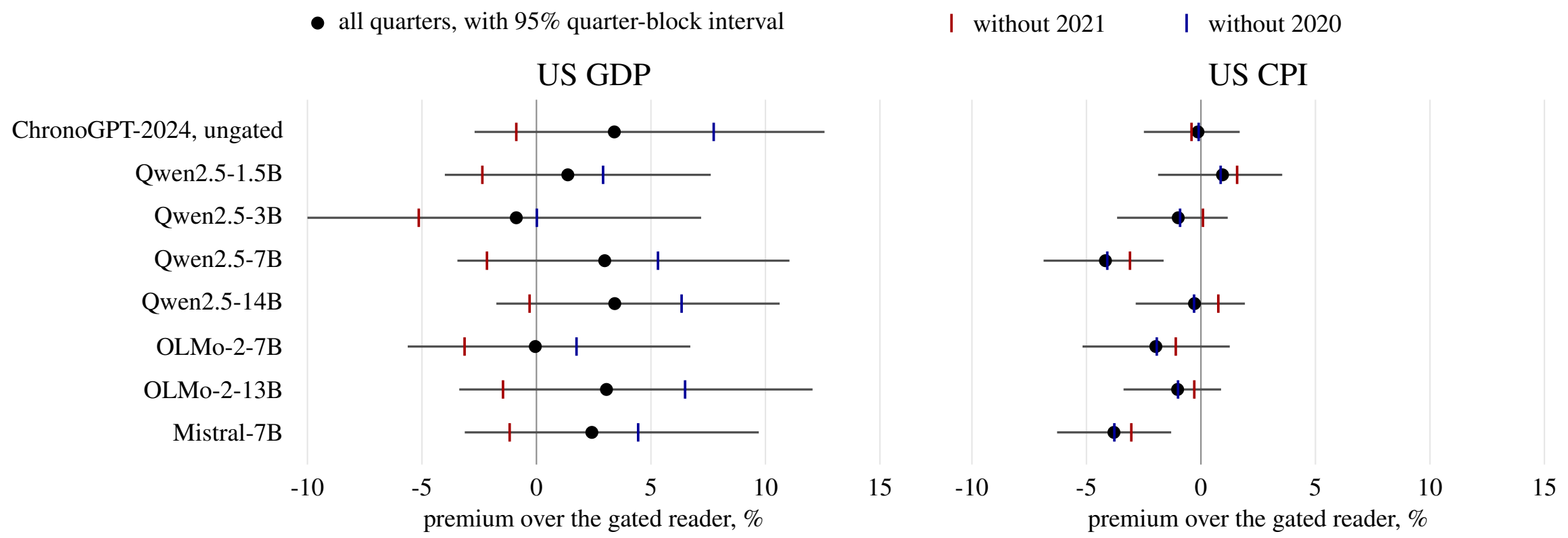

\paragraph{The lookahead premium}
Against the gated reader, the GDP premiums of the \nPremAllGDP{} leaky readers run from
\premAllMinGDP{}\% to \premAllMaxGDP{}\% (Figure~\ref{fig:forest}); all \nCiZeroGDP{}
intervals include zero and the smallest Holm-adjusted $p$ is \holmPremMinGDP{}. The
premiums depend on two years. Without 2021 they run from \premXMinGDP{}\% to
\premXMaxGDP{}\%; without 2020, which contributes 2020\,Q1 and Q4, from
\premXXMinGDP{}\% to \premXXMaxGDP{}\%. Every one of these lies below the minimum
detectable effect of \mdeMinGDP{}\% to \mdeMaxGDP{}\%. The 2021 swing is not a mark of
memory: ChronoGPT with a 2011 cutoff, which cannot know any test-window outcome, also
loses its premium without 2021 (\prxFixedAGDP{}\%), which points to a poor 2021 for the
gated reader rather than to knowledge of 2021 in the others (Figure~\ref{fig:paths}, Appendix~\ref{app:figs}).
On CPI the premiums run from \premAllMinCPI{}\% to \premAllMaxCPI{}\%, and two leaky
readers are worse than the gated one with intervals that exclude zero: Qwen2.5-7B
(\prQwenSevenCPI{}\%, $[\prloQwenSevenCPI{}, \prhiQwenSevenCPI{}]$, Coroneo--Iacone
$p=\prciQwenSevenCPI{}$, Harvey--Leybourne--Newbold $p=\prhlnQwenSevenCPI{}$) and
Mistral-7B (\prMistralSevenCPI{}\%, $p=\prciMistralSevenCPI{}$ and \prhlnMistralSevenCPI{});
after Holm adjustment within CPI the smallest $p$ is \holmPremMinCPI{}.

The output gate moves accuracy by about as much as the premiums. Against its unguarded
signal (which also skips the calibration), the gated reader's guarded signal changes GDP
accuracy by \guardgGatedGDP{}\%; with unguarded signals throughout, the GDP premiums run
from \premRawMinGDP{}\% to \premRawMaxGDP{}\%. The rotation placebo finds little: against the bridge, the real ChronoGPT-2024 signal beats \plpctFixed{}\% of its
\nRotations{} rotations (a one-sided $p$ of about 0.08) and the real gated signal
\plpctGated{}\% of its own; measured against the real gated reader, rotated
ChronoGPT-2024 signals, misaligned with every outcome, reach the observed premium in
\plpremShare{}\% of rotations.

\begin{table*}[t]
\centering
\caption{Real-time GDP nowcasts, 2012--2024, \nOrigins{} origins over \nQuarters{}
quarters. \rmse{} against first release. Premium: percentage reduction in \rmse{}
against the gated ChronoGPT reader's text model, with a 95\% interval from resampling
quarters; the premium without 2021, without 2020, and its range when each year is left
out in turn; and the premium with unguarded signals for both readers. CPI in
Table~\ref{tab:valuecpi}.}
\label{tab:value}
\fitwidth{\scriptsize\begin{tabular}{@{}l r l r r l l@{}}
\toprule
Reader & \rmse{} & premium, \% [95\%] & ex-2021 & ex-2020 & each year left out & unguarded premium [95\%] \\
\midrule
No text (factor bridge) & 1.913 & & & & & \\
ChronoGPT, gated & 1.962 & ref. & & & & ref. \\
ChronoGPT, cutoff 2011 & 1.951 & $+0.6$ $[-6.4, +9.7]$ & $-4.1$ & $+4.9$ & $[-4.1, +4.9]$ & $-0.7$ $[-6.0, +6.6]$ \\
ChronoGPT, cutoff 2015 & 1.977 & $-0.8$ $[-7.1, +7.3]$ & $-3.0$ & $+3.5$ & $[-3.0, +3.5]$ & $-1.8$ $[-6.5, +5.3]$ \\
ChronoGPT, cutoff 2019 & 1.965 & $-0.1$ $[-6.4, +8.2]$ & $-3.2$ & $+3.7$ & $[-3.2, +3.7]$ & $-1.5$ $[-6.7, +5.9]$ \\
ChronoGPT, cutoff 2024 & 1.895 & $+3.4$ $[-2.7, +12.6]$ & $-0.9$ & $+7.7$ & $[-0.9, +7.7]$ & $+1.3$ $[-3.5, +8.2]$ \\
\addlinespace
Qwen2.5-1.5B & 1.935 & $+1.4$ $[-4.0, +7.6]$ & $-2.4$ & $+2.9$ & $[-2.4, +2.9]$ & $-0.3$ $[-4.5, +5.3]$ \\
Qwen2.5-3B & 1.979 & $-0.9$ $[-10.1, +7.2]$ & $-5.1$ & $+0.0$ & $[-5.1, +1.6]$ & $+1.1$ $[-3.1, +6.7]$ \\
Qwen2.5-7B & 1.903 & $+3.0$ $[-3.5, +11.1]$ & $-2.2$ & $+5.3$ & $[-2.2, +5.3]$ & $+0.2$ $[-4.7, +6.3]$ \\
Qwen2.5-14B & 1.895 & $+3.4$ $[-1.8, +10.6]$ & $-0.3$ & $+6.3$ & $[-0.3, +6.3]$ & $+1.1$ $[-3.6, +8.2]$ \\
OLMo-2-7B & 1.963 & $-0.1$ $[-5.6, +6.7]$ & $-3.1$ & $+1.8$ & $[-3.1, +1.8]$ & $-1.8$ $[-7.3, +5.9]$ \\
OLMo-2-13B & 1.902 & $+3.1$ $[-3.4, +12.1]$ & $-1.5$ & $+6.5$ & $[-1.5, +6.5]$ & $-0.7$ $[-8.4, +7.1]$ \\
Mistral-7B & 1.914 & $+2.4$ $[-3.1, +9.7]$ & $-1.2$ & $+4.4$ & $[-1.2, +4.4]$ & $-3.0$ $[-12.8, +5.5]$ \\
Dictionary & 1.907 & n/a & & & & n/a \\
\bottomrule
\end{tabular}
}
\end{table*}

\paragraph{Professional nowcasts}
No reader's signal corrects the SPF, GDPNow or New York Fed nowcasts beyond what
removing their mean error does: across the \encN{} tests of a text correction against a
mean-error correction, the smallest Holm-adjusted $p$ is \encMinHolm{}
(Table~\ref{tab:encompass}). Against GDPNow the mean-error correction alone improves the
nowcast.

\section{Discussion}\label{sec:discussion}

\paragraph{What the evidence supports}
The larger modern open models hold outcome memory that a date retrieves, beyond what
memory-free readers achieve on the same items. Whether the small chronologically
consistent models hold any is not resolvable with these probes. In a real-time nowcast,
neither the memory nor the text shows up as resolvable skill. A probe can show that a
model could use the future, but only a forecasting comparison can show that it does, and
at macroeconomic sample sizes that comparison has little power.

\paragraph{Why the forecast comparison has little power}
With \nOrigins{} origins over \nQuarters{} quarters, the primary test detects a
premium of about \mdeMinGDP{} to \mdeMaxGDP{} percent of GDP nowcast \rmse{} with 80\%
power, larger than every premium we estimate with or without 2020 and 2021. The
swings of 2020 and 2021 move the sample \rmse{} more than all other quarters together,
which is why the premium changes sign depending on which of the two years is left out.
Audits of language-model forecasters that rest on forecast accuracy alone should
report the minimum detectable effect and the leave-one-period-out premium.

\paragraph{Beyond central-bank text}
The design needs only dated inputs, outcomes with known release dates, and one family of
checkpoints with known cutoffs, as in forecasting benchmarks of questions that resolve on
known dates or problems sorted by release date \citep{roberts2024cutoff}.

\paragraph{Limitations}
The documents are one central bank's, and the probes test two outcomes. The
chronologically consistent families available are small, so the contrast between a
chronologically consistent reader and a modern one confounds the cutoff with scale
and training volume; the ladder within each family does not, but it measures
small models. Training cutoffs are documented only for OLMo-2's base model; for the other modern
readers we bound them by release dates. Items after a reader's cutoff are mislabelled as leaky, which biases its probe AUC
toward chance, not away from it.
The gate is conservative: a document of year $y$ is read by the checkpoint that
ends in $y-1$, so the gated reader misses up to a year of legitimate context.
The nowcaster's fill was chosen on a hold-out
after two other fills had been seen on the test window; all six specifications are
reported.

\section*{Use of AI assistants}
The authors used AI assistants for editing text and for coding, under their direction,
and take full responsibility for the content.

\bibliography{refs}

\appendix
\setcounter{dbltopnumber}{3}
\renewcommand{\dbltopfraction}{0.95}\renewcommand{\dblfloatpagefraction}{0.8}

\section{Replication details}\label{app:repl}

\subsection{Readers}\label{app:readers}
Table~\ref{tab:readers} gives each reader's size, the end of its training text and its role.

\begin{table*}[t]
\centering
\caption{Readers. Release dates are the public Hugging Face release of the
checkpoint used. For OLMo-2 we measure in its indexed stage-1 pretraining corpus when
FOMC text stops appearing (Section~\ref{sec:membership}); the Instruct model card states
no cutoff.}
\label{tab:readers}
\footnotesize
\fitwidth{\begin{tabular}{@{}l l l l@{}}
\toprule
Reader & Parameters & Training text ends & Role \\
\midrule
ChronoGPT-Instruct, 26 checkpoints & 1.55B & 31 Dec of each year, 1999--2024 & gated \\
ChronoGPT-Instruct, fixed checkpoint & 1.55B & 31 Dec 2011, 2015, 2019, 2024 & ungated ladder \\
DatedGPT-Instruct, 12 checkpoints & 1.34B & 31 Dec of each year, 2013--2024 & gated, probes only \\
DatedGPT-Instruct, fixed checkpoint & 1.34B & 31 Dec 2016, 2020, 2024 & ungated ladder, probes only \\
Qwen2.5-Instruct \citep{qwen2025report} & 1.5B, 3B, 7B, 14B & undocumented; released Sep 2024 & ungated \\
OLMo-2-1124-Instruct \citep{olmo2025} & 7B, 13B & Dec 2023 (base model card); FOMC text to \igLast{} & ungated \\
Mistral-7B-Instruct-v0.3 \citep{jiang2023mistral} & 7B & undocumented; released May 2024 & ungated \\
Dictionary of policy phrases & none & phrases written after the sample & reference \\
\bottomrule
\end{tabular}}
\end{table*}

\subsection{Prompts}\label{app:prompts}
Every document question uses the template below; the questions and probe
prompts are listed in Table~\ref{tab:prompts}, read from the scoring code. Chat
models receive the prompt as one user turn through their own chat template. The
ChronoGPT-Instruct readers receive it in the authors' instruction format, and the
DatedGPT-Instruct readers through their chat template, encoded without a second
beginning-of-sequence token.
\begin{quote}\small\ttfamily
Text from a Federal Open Market Committee document:\\
"""\\
\{excerpt\}\\
"""\\[3pt]
\{question\} Answer with one word: \{answer 1\} or \{answer 2\}.
\end{quote}
The content-free input used for calibration replaces the excerpt with ``N/A''. The
probe prompts end with the same instruction to answer in one word.

\begin{table*}[t]
\centering
\caption{Every question asked of every reader. Three wordings per document
dimension; one per probe. The answer pair is fixed per dimension.}
\label{tab:prompts}
\scriptsize
\fitwidth{\begin{tabular}{@{}p{2.3cm}p{1.6cm}p{10.2cm}@{}}
\toprule
Dimension & Answers & Question \\
\midrule
Stance & more / less & Compared with a neutral stance, is the monetary policy described in this text more or less restrictive? \\
 &  & Is the Committee in this text setting policy that is more or less restrictive than neutral? \\
 &  & Overall, does this text describe monetary policy that is more or less restrictive? \\
\addlinespace
Inflation risk & high / low & Are the risks of inflation rising described in this text high or low? \\
 &  & How large is the concern about inflation pressure in this text, high or low? \\
 &  & Is the upside risk to inflation in this text high or low? \\
\addlinespace
Growth risk & high / low & Are the downside risks to economic growth in this text high or low? \\
 &  & How large is the concern about weak economic growth in this text, high or low? \\
 &  & Is the risk of an economic slowdown in this text high or low? \\
\addlinespace
Uncertainty & high / low & Is the economic uncertainty described in this text high or low? \\
 &  & How uncertain is the outlook described in this text, high or low? \\
 &  & Is the level of uncertainty expressed in this text high or low? \\
\addlinespace
Forward guidance & higher / lower & Does the guidance about future policy in this text point to higher or lower interest rates ahead? \\
 &  & Is the expected path of the policy rate in this text higher or lower? \\
 &  & Does this text signal higher or lower interest rates in the coming months? \\
\addlinespace
\midrule
Decision probe & raise / lower & On <date>, the Federal Open Market Committee changed its target for the federal funds rate. Did it raise or lower the target? \\
GDP probe & above / below & According to the first official estimate, was US real GDP growth above or below 2 percent at an annual rate in <date>? \\
\bottomrule
\end{tabular}
}
\end{table*}

\begin{table*}[t]
\centering
\caption{All six nowcaster specifications without text, on the hold-out window used
for selection and on the test window. Starred: the selected specification.}
\label{tab:specs}
\small
\fitwidth{\begin{tabular}{@{}ll rr rr@{}}
\toprule
& & \multicolumn{2}{c}{Hold-out 2006--2011} & \multicolumn{2}{c}{Test 2012--2024} \\
\cmidrule(lr){3-4}\cmidrule(lr){5-6}
Ragged-edge fill & Indicators & GDP & CPI & GDP & CPI \\
\midrule
mean growth & all eleven & 1.831 & 2.545 & 2.376 & 1.675 \\
mean growth & by target & 1.704 & 2.303 & 2.419 & 1.327 \\
no change & all eleven & 1.896 & 2.375 & 1.921 & 1.577 \\
no change & by target & 1.751 & 2.089 & 1.933 & 1.655 \\
AR(2) & all eleven & 1.926 & 2.200 & 1.903 & 1.281 \\
AR(2)$^{*}$ & by target & 1.798 & 1.772 & 1.913 & 1.148 \\
\bottomrule
\end{tabular}
}
\end{table*}

\subsection{Corpus construction and cleaning}\label{app:corpus}
The corpus is built from the Federal Reserve's own index of documents and release
dates. Four corrections were applied before any result reported here.
\begin{enumerate}
\item \textbf{Misdated minutes.} Twenty-eight raw files held copies of minutes stored
under the date of an intermeeting conference call, between three and sixty-six
days before the minutes were published, which placed text before its public release
for two test origins and thirteen training cells. Files with no entry in the Federal
Reserve's index are not used.
\item \textbf{Missing minutes.} Twenty-four sets of minutes from 2004--2006 were
stored under non-standard file names. They are included, dated by their release.
\item \textbf{Excerpt boundaries and dates.} Twenty-six statements without a
recorded vote run into the web page's navigation text; their excerpts stop at the page
footer. Eight statements carried a date one day before their release; statements are
dated by their press-release URL. We added section markers for minutes whose outlook
or policy discussion the standard markers miss; \nOneSectionMinutes{} sets of minutes
still yield only one of the two sections.
\item \textbf{Character encoding.} Sixty-eight excerpts carried UTF-8 punctuation
decoded as Latin-1. They are decoded as UTF-8.
\end{enumerate}
Labels for the decision probe come from the Federal Reserve's published table of
target changes, \nProbeDecision{} changes from 2000 to 2025, each dated by the statement
that announced it. Every reader scored the corrected documents and probe items.

\subsection{Determinism}\label{app:repro}
Scores are deterministic up to floating-point noise. The ungated ChronoGPT-2024 reader
was scored twice, on an Apple-silicon GPU and on a CUDA GPU, before the corpus cleaning;
the \xcN{} raw scores differ by at most $\xcMaxDiff{}$ and agree in sign in
\xcSignAgree{}\% of cases. Items changed by the cleaning were then scored again.

\subsection{Nowcaster specifications}\label{app:specs}
Six specifications were evaluated: three ways of filling the months not yet
published (the mean growth of the last three months, no change, and an
autoregression of order two on each indicator's own monthly changes), each with all
eleven indicators or with indicator blocks by target. The fills were run in the order
listed, and once the autoregressive fill was added the selection rule (the single specification with the lowest mean hold-out \rmse{}
across the two targets, on origins 2006--2011) was applied to all six. Four of the six
had been seen on the test window before the rule was set. Table~\ref{tab:specs}
reports all six on both windows and Table~\ref{tab:robust} the gated and
ChronoGPT-2024 readers under each.

\subsection{Baselines}\label{app:baselines}
Professional nowcasts are used only where their value was public on its date. The
dynamic factor model uses the same vintages, blocks and training quarters as the
bridge; its number of factors and their order were chosen on the hold-out (two factors
with two lags for GDP, two with one lag for CPI). For Table~\ref{tab:encompass}, the
professional's first-release error is regressed, out of sample on an expanding window,
on the latest document's signal, and the corrected nowcast is compared with the
professional one by the Clark--West test and with a correction for the professional's
mean error alone.

\section{Supplementary tables and figures}\label{app:figs}
\label{app:probes}\label{app:gate}\label{app:membership}
Tables and figures referenced from the main text.

\begin{table*}[p]
\centering
\caption{GDP probe within years over all items (raw, and after removing each reader's
mean score per quarter word, with permutation tests within years), the GDP probe AUC on the
2012--2024 items alone, and the decision probe with its circular-shift $p$. Before the
quarter-word adjustment, the ChronoGPT checkpoints with 2015, 2019 and 2024 cutoffs score
\pgyrawFixedB{}, \pgyrawFixedC{} and \pgyrawFixed{} within years; after it,
\pgyFixedB{}, \pgyFixedC{} and \pgyFixed{}. On the decision probe the gated DatedGPT reader
scores \pdDatedGated{} on its \pdnDatedGated{} items and ChronoGPT with a 2011 cutoff
\pdFixedA{}.}
\label{tab:probesapp}
\fitwidth{\scriptsize\begin{tabular}{@{}l rr rr r rrr@{}}
\toprule
& \multicolumn{2}{c}{Same year} & \multicolumn{2}{c}{Same year, word-adjusted} & 2012--2024 & \multicolumn{3}{c}{Decision probe} \\
\cmidrule(lr){2-3}\cmidrule(lr){4-5}\cmidrule(lr){6-6}\cmidrule(lr){7-9}
Reader & AUC & $p$ & AUC & $p$ & AUC & $n$ & AUC & circular $p$ \\
\midrule
ChronoGPT, gated & 0.48 & 0.612 & 0.42 & 0.854 & 0.53 & 74 & 0.50 & 0.554 \\
ChronoGPT, cutoff 2011 & 0.58 & 0.206 & 0.47 & 0.678 & 0.60 & 74 & 0.65 & 0.041 \\
ChronoGPT, cutoff 2015 & 0.70 & 0.006 & 0.52 & 0.469 & 0.64 & 74 & 0.44 & 0.743 \\
ChronoGPT, cutoff 2019 & 0.73 & 0.002 & 0.61 & 0.107 & 0.48 & 74 & 0.54 & 0.297 \\
ChronoGPT, cutoff 2024 & 0.72 & 0.004 & 0.53 & 0.380 & 0.62 & 74 & 0.46 & 0.743 \\
DatedGPT, gated & 0.48 & 0.613 & 0.48 & 0.621 & 0.54 & 31 & 0.79 & 0.065 \\
DatedGPT, cutoff 2016 & 0.64 & 0.057 & 0.52 & 0.471 & 0.65 & 74 & 0.61 & 0.041 \\
DatedGPT, cutoff 2020 & 0.27 & 0.999 & 0.44 & 0.806 & 0.33 & 74 & 0.64 & 0.081 \\
DatedGPT, cutoff 2024 & 0.44 & 0.814 & 0.62 & 0.072 & 0.47 & 74 & 0.49 & 0.540 \\
Qwen2.5-1.5B & 0.48 & 0.595 & 0.62 & 0.075 & 0.56 & 74 & 0.64 & 0.243 \\
Qwen2.5-3B & 0.66 & 0.035 & 0.62 & 0.080 & 0.70 & 74 & 0.67 & 0.054 \\
Qwen2.5-7B & 0.58 & 0.175 & 0.59 & 0.145 & 0.71 & 74 & 0.70 & 0.013 \\
Qwen2.5-14B & 0.66 & 0.030 & 0.61 & 0.114 & 0.73 & 74 & 0.95 & 0.013 \\
OLMo-2-7B & 0.66 & 0.030 & 0.53 & 0.386 & 0.70 & 74 & 0.73 & 0.054 \\
OLMo-2-13B & 0.56 & 0.244 & 0.56 & 0.248 & 0.68 & 74 & 0.86 & 0.013 \\
Mistral-7B & 0.66 & 0.030 & 0.61 & 0.092 & 0.77 & 74 & 0.77 & 0.013 \\
Persistence rule & 0.44 & 0.897 & 0.41 & 0.961 & 0.47 & 74 & 0.54 & 0.392 \\
\bottomrule
\end{tabular}
}
\end{table*}

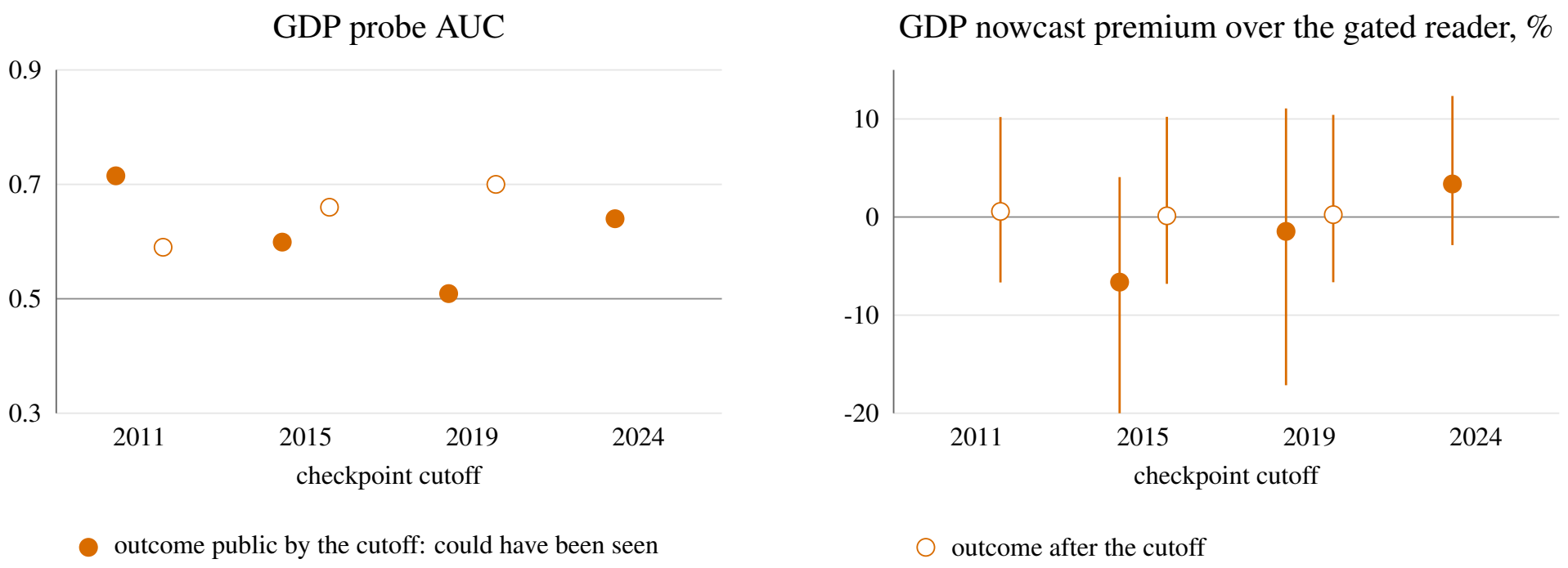
\begin{figure*}[p]
\centering
\resizebox{0.8\textwidth}{!}{\begin{tikzpicture}[font=\scriptsize]
\draw[black!10] (0.000,0.000) -- (6.200,0.000); \node[left] at (0.000,0.000) {0.3};
\draw[black!45] (0.000,1.067) -- (6.200,1.067); \node[left] at (0.000,1.067) {0.5};
\draw[black!10] (0.000,2.133) -- (6.200,2.133); \node[left] at (0.000,2.133) {0.7};
\draw[black!10] (0.000,3.200) -- (6.200,3.200); \node[left] at (0.000,3.200) {0.9};
\draw[black!60] (0.000,0) -- (0.000,3.200);
\node[font=\footnotesize,anchor=south] at (3.100,3.300) {GDP probe AUC};
\node at (3.100,-0.6) {checkpoint cutoff};
\node[below] at (0.775,0) {2011};
\draw[orange!85!black,fill=orange!85!black] (0.555,2.213) circle (2.3pt);
\draw[orange!85!black,fill=white] (0.995,1.547) circle (2.3pt);
\node[below] at (2.325,0) {2015};
\draw[orange!85!black,fill=orange!85!black] (2.105,1.595) circle (2.3pt);
\draw[orange!85!black,fill=white] (2.545,1.920) circle (2.3pt);
\node[below] at (3.875,0) {2019};
\draw[orange!85!black,fill=orange!85!black] (3.655,1.115) circle (2.3pt);
\draw[orange!85!black,fill=white] (4.095,2.133) circle (2.3pt);
\node[below] at (5.425,0) {2024};
\draw[orange!85!black,fill=orange!85!black] (5.205,1.813) circle (2.3pt);
\draw[black!10] (7.800,0.000) -- (14.000,0.000); \node[left] at (7.800,0.000) {-20};
\draw[black!10] (7.800,0.914) -- (14.000,0.914); \node[left] at (7.800,0.914) {-10};
\draw[black!45] (7.800,1.829) -- (14.000,1.829); \node[left] at (7.800,1.829) {0};
\draw[black!10] (7.800,2.743) -- (14.000,2.743); \node[left] at (7.800,2.743) {10};
\draw[black!60] (7.800,0) -- (7.800,3.200);
\node[font=\footnotesize,anchor=south] at (10.900,3.300) {GDP nowcast premium over the gated reader, \%};
\node at (10.900,-0.6) {checkpoint cutoff};
\node[below] at (8.575,0) {2011};
\draw[orange!85!black,line width=0.6pt] (8.795,1.219) -- (8.795,2.761);
\draw[orange!85!black,fill=white] (8.795,1.881) circle (2.3pt);
\node[below] at (10.125,0) {2015};
\draw[orange!85!black,line width=0.6pt] (9.905,0.000) -- (9.905,2.201);
\draw[orange!85!black,fill=orange!85!black] (9.905,1.222) circle (2.3pt);
\draw[orange!85!black,line width=0.6pt] (10.345,1.206) -- (10.345,2.763);
\draw[orange!85!black,fill=white] (10.345,1.840) circle (2.3pt);
\node[below] at (11.675,0) {2019};
\draw[orange!85!black,line width=0.6pt] (11.455,0.261) -- (11.455,2.841);
\draw[orange!85!black,fill=orange!85!black] (11.455,1.695) circle (2.3pt);
\draw[orange!85!black,line width=0.6pt] (11.895,1.221) -- (11.895,2.781);
\draw[orange!85!black,fill=white] (11.895,1.851) circle (2.3pt);
\node[below] at (13.225,0) {2024};
\draw[orange!85!black,line width=0.6pt] (13.005,1.567) -- (13.005,2.957);
\draw[orange!85!black,fill=orange!85!black] (13.005,2.137) circle (2.3pt);
\draw[orange!85!black,fill=orange!85!black] (0.3,-1.25) circle (2.3pt) node[right=3pt,black] {outcome public by the cutoff: could have been seen};
\draw[orange!85!black,fill=white] (8.100,-1.25) circle (2.3pt) node[right=3pt,black] {outcome after the cutoff};
\end{tikzpicture}}
\caption{The ChronoGPT ladder: each checkpoint on outcomes it could have seen (filled)
and could not (open), GDP probe AUC (left) and GDP nowcast premium over the gated
reader with 95\% quarter-block intervals (right). Memory predicts filled above open;
no difference is resolvable.}
\label{fig:ladder}
\end{figure*}

\begin{table*}[p]
\centering
\caption{Same items, consecutive checkpoints: GDP probe items whose outcome was released
after the earlier checkpoint's cutoff and by the later one's. Memory predicts the later
checkpoint does better. Intervals from resampling years.}
\label{tab:contrasts}
\fitwidth{\scriptsize\begin{tabular}{@{}l l r rr l@{}}
\toprule
Earlier checkpoint & Later checkpoint & items & earlier AUC & later AUC & later minus earlier [95\%] \\
\midrule
ChronoGPT, cutoff 2011 & ChronoGPT, cutoff 2015 & 16 & 0.47 & 0.62 & $+0.15$ $[-0.20, +0.41]$ \\
ChronoGPT, cutoff 2015 & ChronoGPT, cutoff 2019 & 16 & 0.57 & 0.30 & $-0.27$ $[-0.83, +0.09]$ \\
ChronoGPT, cutoff 2019 & ChronoGPT, cutoff 2024 & 20 & 0.68 & 0.77 & $+0.10$ $[-0.10, +0.51]$ \\
DatedGPT, cutoff 2016 & DatedGPT, cutoff 2020 & 16 & 0.51 & 0.18 & $-0.33$ $[-1.00, +0.08]$ \\
DatedGPT, cutoff 2020 & DatedGPT, cutoff 2024 & 16 & 0.35 & 0.33 & $-0.02$ $[-0.50, +0.33]$ \\
\bottomrule
\end{tabular}
}
\end{table*}

\begin{figure*}[p]
\centering
\resizebox{0.75\textwidth}{!}{\input{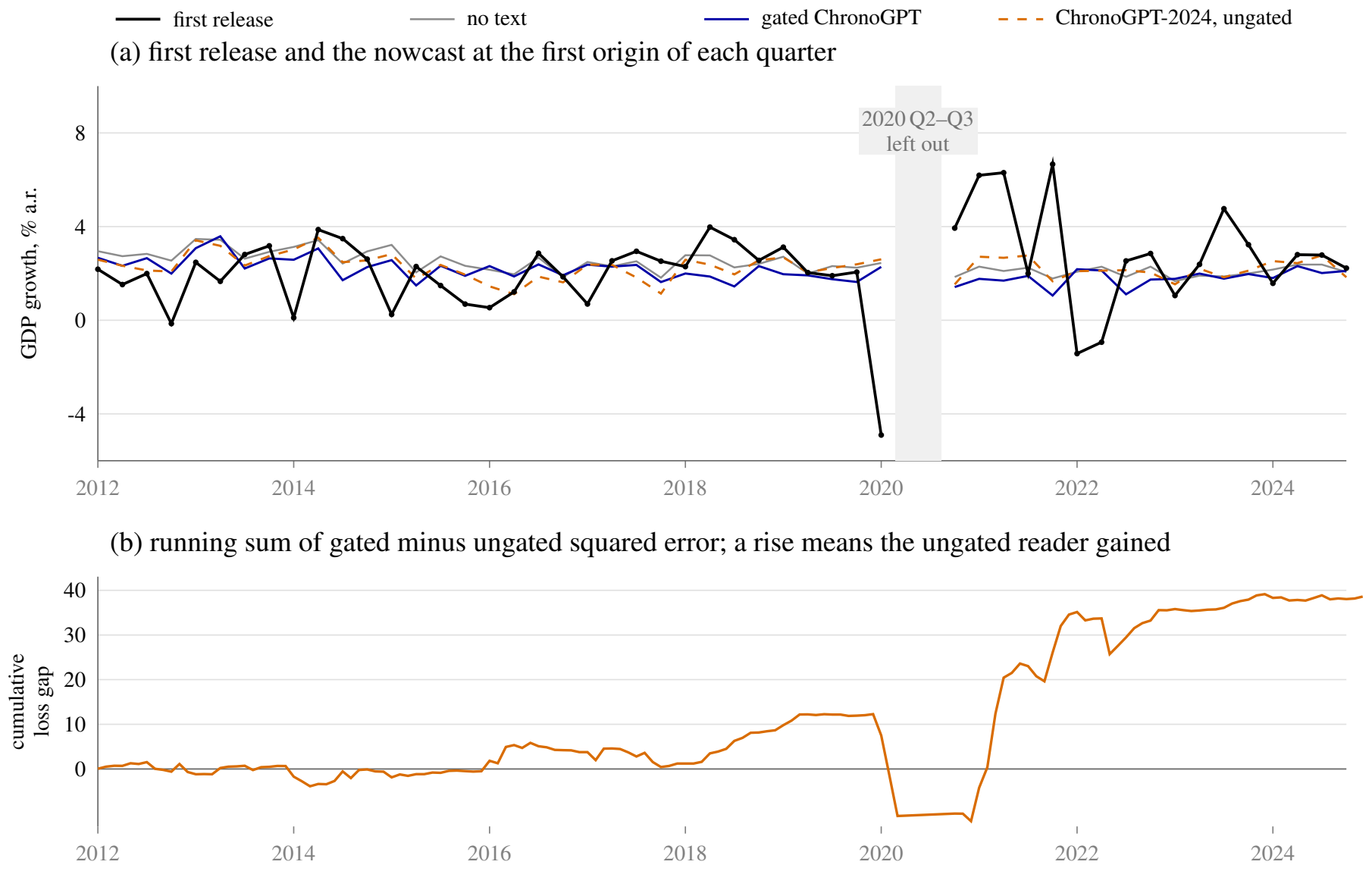}}
\caption{(a) First-release GDP growth and three nowcasts at the first origin of each
quarter. (b) Running sum of the gated reader's squared error minus the ungated
ChronoGPT-2024 reader's: the gap rises before 2020, falls in 2020 and rises in 2021.}
\label{fig:paths}
\end{figure*}

\begin{table*}[p]
\centering
\caption{Real-time CPI nowcasts, as in Table~\ref{tab:value}.}
\label{tab:valuecpi}
\fitwidth{\scriptsize\begin{tabular}{@{}l r l r r l l@{}}
\toprule
Reader & \rmse{} & premium, \% [95\%] & ex-2021 & ex-2020 & each year left out & unguarded premium [95\%] \\
\midrule
No text (factor bridge) & 1.148 & & & & & \\
ChronoGPT, gated & 1.161 & ref. & & & & ref. \\
ChronoGPT, cutoff 2011 & 1.152 & $+0.7$ $[-1.0, +2.6]$ & $+1.2$ & $+0.6$ & $[+0.5, +1.2]$ & $+0.1$ $[-1.8, +2.1]$ \\
ChronoGPT, cutoff 2015 & 1.157 & $+0.3$ $[-1.8, +2.4]$ & $+1.6$ & $+0.4$ & $[-0.2, +1.6]$ & $+0.2$ $[-1.9, +2.1]$ \\
ChronoGPT, cutoff 2019 & 1.156 & $+0.4$ $[-1.5, +2.3]$ & $+1.3$ & $+0.4$ & $[+0.2, +1.3]$ & $-0.1$ $[-2.7, +2.2]$ \\
ChronoGPT, cutoff 2024 & 1.162 & $-0.1$ $[-2.5, +1.7]$ & $-0.4$ & $-0.1$ & $[-0.4, +0.5]$ & $-0.1$ $[-2.3, +1.7]$ \\
\addlinespace
Qwen2.5-1.5B & 1.150 & $+0.9$ $[-1.9, +3.5]$ & $+1.6$ & $+0.9$ & $[+0.5, +1.6]$ & $+1.9$ $[-0.6, +4.6]$ \\
Qwen2.5-3B & 1.172 & $-1.0$ $[-3.7, +1.2]$ & $+0.1$ & $-0.9$ & $[-2.3, +0.1]$ & $+0.4$ $[-2.3, +2.5]$ \\
Qwen2.5-7B & 1.209 & $-4.2$ $[-6.9, -1.6]$ & $-3.1$ & $-4.1$ & $[-6.0, -3.1]$ & $-0.4$ $[-3.5, +2.5]$ \\
Qwen2.5-14B & 1.164 & $-0.3$ $[-2.9, +1.9]$ & $+0.8$ & $-0.3$ & $[-1.3, +0.8]$ & $-1.6$ $[-5.8, +2.3]$ \\
OLMo-2-7B & 1.184 & $-2.0$ $[-5.2, +1.3]$ & $-1.1$ & $-1.9$ & $[-2.4, -1.1]$ & $-0.2$ $[-2.3, +1.9]$ \\
OLMo-2-13B & 1.173 & $-1.0$ $[-3.4, +0.9]$ & $-0.3$ & $-1.0$ & $[-2.0, -0.3]$ & $-2.0$ $[-7.3, +1.4]$ \\
Mistral-7B & 1.205 & $-3.8$ $[-6.3, -1.3]$ & $-3.0$ & $-3.8$ & $[-4.8, -2.9]$ & $-3.7$ $[-7.5, -1.0]$ \\
Dictionary & 1.172 & n/a & & & & n/a \\
\bottomrule
\end{tabular}
}
\end{table*}

\begin{table*}[p]
\centering
\caption{The gated and ChronoGPT-2024 readers under every specification. Premium:
percentage reduction in \rmse{} of the ungated against the gated reader, with the
Coroneo--Iacone $p$. The GDP premium is positive in \robPosOfGDP{} specifications and
significant in \robSigGDP{}; the CPI premium runs from \robMinCPI{}\% to \robMaxCPI{}\%
and is nominally significant in \robSigCPI{}.}
\label{tab:robust}
\scriptsize
\fitwidth{\begin{tabular}{@{}ll rrrr rrrr@{}}
\toprule
& & \multicolumn{4}{c}{US GDP} & \multicolumn{4}{c}{US CPI} \\
\cmidrule(lr){3-6}\cmidrule(lr){7-10}
Fill & Blocks & factor & gated & ungated & premium ($p$) & factor & gated & ungated & premium ($p$) \\
\midrule
mean growth & all eleven & 2.376 & 2.322 & 2.366 & $-1.9$ (0.58) & 1.675 & 1.646 & 1.719 & $-4.5$ (0.26) \\
mean growth & by target & 2.419 & 2.349 & 2.390 & $-1.8$ (0.54) & 1.327 & 1.334 & 1.356 & $-1.6$ (0.25) \\
no change & all eleven & 1.921 & 2.038 & 1.949 & $+4.4$ (0.13) & 1.577 & 1.724 & 1.612 & $+6.5$ (0.01) \\
no change & by target & 1.933 & 2.052 & 1.951 & $+4.9$ (0.15) & 1.655 & 1.673 & 1.672 & $+0.1$ (0.93) \\
AR(2) & all eleven & 1.903 & 1.943 & 1.899 & $+2.3$ (0.37) & 1.281 & 1.235 & 1.314 & $-6.4$ (0.10) \\
AR(2)$^{*}$ & by target & 1.913 & 1.962 & 1.895 & $+3.4$ (0.28) & 1.148 & 1.161 & 1.162 & $-0.1$ (0.92) \\
\bottomrule
\end{tabular}
}
\end{table*}

\begin{table*}[p]
\centering
\caption{Every forecast against first release, 2012--2024, pandemic quarters
excluded, each on its own coverage. The univariate rules make one forecast per quarter, so
the expanding mean differs slightly from the running mean scored at the \nOrigins{}
monthly origins in Section~\ref{sec:results:value}.}
\label{tab:bench}
\scriptsize
\fitwidth{\begin{tabular}{@{}l l rr@{}}
\toprule
Forecast & Origins & US GDP & US CPI \\
\midrule
Survey of Professional Forecasters, median & mid-quarter, 50 quarters & 1.748 & 1.430 \\
GDPNow (Atlanta Fed), public from July 2014 & weekly, 40 quarters & 1.968 & n/a \\
New York Fed Staff Nowcast, live periods & weekly, 26 quarters & 1.752 & n/a \\
AR($p$), BIC & one per quarter & 2.365 & 2.094 \\
Naive last & one per quarter & 4.151 & 1.958 \\
Seasonal naive & one per quarter & 8.066 & 2.995 \\
Expanding mean & one per quarter & 1.900 & 2.399 \\
\midrule
Real-time factor bridge (this paper) & month-end, 150 & 1.913 & 1.148 \\
\bottomrule
\end{tabular}
}
\end{table*}

\begin{table*}[p]
\centering
\caption{Text on top of the professional nowcasts. Bias-only: the professional
nowcast plus its mean past error. Best text reader: the reader whose out-of-sample
correction has the lowest \rmse{}; ratio to the professional \rmse{}. Holm $p$: the
Holm-adjusted Clark--West $p$ of that reader's text correction against the mean-error
correction, over all \encN{} text tests. \rmse{} is over the origins scored out of
sample, so it differs slightly from Table~\ref{tab:bench}, which weights each quarter
once.}
\label{tab:encompass}
\scriptsize
\fitwidth{\begin{tabular}{@{}l l r r r l r r@{}}
\toprule
Nowcast & Target & quarters & \rmse{} & bias-only & best text reader & ratio & Holm $p$ \\
\midrule
GDPNow & GDP & 32 & 1.986 & 0.949 & Qwen2.5-3B & 0.975 & 1.00 \\
New York Fed & GDP & 19 & 1.945 & 1.031 & ChronoGPT, cutoff 2019 & 1.078 & 1.00 \\
SPF & CPI & 50 & 1.437 & 1.002 & Mistral-7B & 0.944 & 1.00 \\
SPF & GDP & 50 & 1.748 & 1.001 & ChronoGPT, cutoff 2015 & 0.996 & 1.00 \\
\bottomrule
\end{tabular}
}
\end{table*}

\begin{figure*}[p]
\centering
\resizebox{0.7\textwidth}{!}{\begin{tikzpicture}[font=\scriptsize]
\node[anchor=south,align=center] at (0.725,3.830) {stance};
\node[anchor=south,align=center] at (2.175,3.830) {inflation\\risk};
\node[anchor=south,align=center] at (3.625,3.830) {growth\\risk};
\node[anchor=south,align=center] at (5.075,3.830) {uncer-\\tainty};
\node[anchor=south,align=center] at (6.525,3.830) {forward\\guidance};
\node[anchor=east] at (-0.08,3.570) {ChronoGPT, vintage-gated};
\fill[blue!15!white] (0.000,3.360) rectangle ++(1.45,0.42);
\node[black] at (0.725,3.570) {9\%};
\fill[blue!2!white] (1.450,3.360) rectangle ++(1.45,0.42);
\node[black] at (2.175,3.570) {1\%};
\fill[blue!1!white] (2.900,3.360) rectangle ++(1.45,0.42);
\node[black] at (3.625,3.570) {1\%};
\fill[blue!6!white] (4.350,3.360) rectangle ++(1.45,0.42);
\node[black] at (5.075,3.570) {4\%};
\fill[blue!58!white] (5.800,3.360) rectangle ++(1.45,0.42);
\node[white] at (6.525,3.570) {35\%};
\node[anchor=east] at (-0.08,3.150) {ChronoGPT-2024, ungated};
\fill[blue!5!white] (0.000,2.940) rectangle ++(1.45,0.42);
\node[black] at (0.725,3.150) {3\%};
\fill[blue!8!white] (1.450,2.940) rectangle ++(1.45,0.42);
\node[black] at (2.175,3.150) {5\%};
\fill[blue!3!white] (2.900,2.940) rectangle ++(1.45,0.42);
\node[black] at (3.625,3.150) {2\%};
\fill[blue!24!white] (4.350,2.940) rectangle ++(1.45,0.42);
\node[black] at (5.075,3.150) {15\%};
\fill[blue!47!white] (5.800,2.940) rectangle ++(1.45,0.42);
\node[black] at (6.525,3.150) {28\%};
\node[anchor=east] at (-0.08,2.730) {Qwen2.5-1.5B};
\fill[blue!38!white] (0.000,2.520) rectangle ++(1.45,0.42);
\node[black] at (0.725,2.730) {23\%};
\fill[blue!37!white] (1.450,2.520) rectangle ++(1.45,0.42);
\node[black] at (2.175,2.730) {22\%};
\fill[blue!23!white] (2.900,2.520) rectangle ++(1.45,0.42);
\node[black] at (3.625,2.730) {14\%};
\fill[blue!50!white] (4.350,2.520) rectangle ++(1.45,0.42);
\node[black] at (5.075,2.730) {30\%};
\fill[blue!26!white] (5.800,2.520) rectangle ++(1.45,0.42);
\node[black] at (6.525,2.730) {16\%};
\node[anchor=east] at (-0.08,2.310) {Qwen2.5-3B};
\fill[blue!1!white] (0.000,2.100) rectangle ++(1.45,0.42);
\node[black] at (0.725,2.310) {0\%};
\fill[blue!28!white] (1.450,2.100) rectangle ++(1.45,0.42);
\node[black] at (2.175,2.310) {17\%};
\fill[blue!69!white] (2.900,2.100) rectangle ++(1.45,0.42);
\node[white] at (3.625,2.310) {41\%};
\fill[blue!19!white] (4.350,2.100) rectangle ++(1.45,0.42);
\node[black] at (5.075,2.310) {11\%};
\fill[blue!57!white] (5.800,2.100) rectangle ++(1.45,0.42);
\node[white] at (6.525,2.310) {34\%};
\node[anchor=east] at (-0.08,1.890) {Qwen2.5-7B};
\fill[blue!73!white] (0.000,1.680) rectangle ++(1.45,0.42);
\node[white] at (0.725,1.890) {44\%};
\fill[blue!27!white] (1.450,1.680) rectangle ++(1.45,0.42);
\node[black] at (2.175,1.890) {16\%};
\fill[blue!42!white] (2.900,1.680) rectangle ++(1.45,0.42);
\node[black] at (3.625,1.890) {25\%};
\fill[blue!24!white] (4.350,1.680) rectangle ++(1.45,0.42);
\node[black] at (5.075,1.890) {15\%};
\fill[blue!20!white] (5.800,1.680) rectangle ++(1.45,0.42);
\node[black] at (6.525,1.890) {12\%};
\node[anchor=east] at (-0.08,1.470) {Qwen2.5-14B};
\fill[blue!38!white] (0.000,1.260) rectangle ++(1.45,0.42);
\node[black] at (0.725,1.470) {23\%};
\fill[blue!16!white] (1.450,1.260) rectangle ++(1.45,0.42);
\node[black] at (2.175,1.470) {9\%};
\fill[blue!59!white] (2.900,1.260) rectangle ++(1.45,0.42);
\node[white] at (3.625,1.470) {36\%};
\fill[blue!61!white] (4.350,1.260) rectangle ++(1.45,0.42);
\node[white] at (5.075,1.470) {37\%};
\fill[blue!18!white] (5.800,1.260) rectangle ++(1.45,0.42);
\node[black] at (6.525,1.470) {11\%};
\node[anchor=east] at (-0.08,1.050) {OLMo-2-7B};
\fill[blue!84!white] (0.000,0.840) rectangle ++(1.45,0.42);
\node[white] at (0.725,1.050) {51\%};
\fill[blue!57!white] (1.450,0.840) rectangle ++(1.45,0.42);
\node[white] at (2.175,1.050) {34\%};
\fill[blue!85!white] (2.900,0.840) rectangle ++(1.45,0.42);
\node[white] at (3.625,1.050) {51\%};
\fill[blue!12!white] (4.350,0.840) rectangle ++(1.45,0.42);
\node[black] at (5.075,1.050) {7\%};
\fill[blue!30!white] (5.800,0.840) rectangle ++(1.45,0.42);
\node[black] at (6.525,1.050) {18\%};
\node[anchor=east] at (-0.08,0.630) {OLMo-2-13B};
\fill[blue!73!white] (0.000,0.420) rectangle ++(1.45,0.42);
\node[white] at (0.725,0.630) {44\%};
\fill[blue!24!white] (1.450,0.420) rectangle ++(1.45,0.42);
\node[black] at (2.175,0.630) {14\%};
\fill[blue!38!white] (2.900,0.420) rectangle ++(1.45,0.42);
\node[black] at (3.625,0.630) {23\%};
\fill[blue!20!white] (4.350,0.420) rectangle ++(1.45,0.42);
\node[black] at (5.075,0.630) {12\%};
\fill[blue!16!white] (5.800,0.420) rectangle ++(1.45,0.42);
\node[black] at (6.525,0.630) {9\%};
\node[anchor=east] at (-0.08,0.210) {Mistral-7B};
\fill[blue!42!white] (0.000,0.000) rectangle ++(1.45,0.42);
\node[black] at (0.725,0.210) {25\%};
\fill[blue!22!white] (1.450,0.000) rectangle ++(1.45,0.42);
\node[black] at (2.175,0.210) {13\%};
\fill[blue!31!white] (2.900,0.000) rectangle ++(1.45,0.42);
\node[black] at (3.625,0.210) {19\%};
\fill[blue!24!white] (4.350,0.000) rectangle ++(1.45,0.42);
\node[black] at (5.075,0.210) {14\%};
\fill[blue!21!white] (5.800,0.000) rectangle ++(1.45,0.42);
\node[black] at (6.525,0.210) {12\%};
\draw[black!40] (0,0) rectangle (7.25,3.78);
\end{tikzpicture}}
\caption{Share of document-dimension readings set to zero because the three wordings
of the question disagreed in sign. Against its unguarded signal, the gated reader's
guarded signal changes GDP accuracy by \guardgGatedGDP{}\%; with unguarded signals
throughout, the GDP premiums run from \premRawMinGDP{}\% to \premRawMaxGDP{}\% instead of
\premAllMinGDP{}\% to \premAllMaxGDP{}\% (Table~\ref{tab:value}).}
\label{fig:abstain}
\end{figure*}

\begin{table*}[p]
\centering
\caption{Membership scores ($k=20$). ChronoGPT rows: AUC of each score for documents
released before the checkpoint's cutoff (possible members) against documents released
after it (certain non-members). OLMo rows: AUC against presence of a document's novel
text in OLMo-2's indexed stage-1 corpus, over all documents with a novel span.}
\label{tab:membership}
\scriptsize
\fitwidth{\begin{tabular}{@{}l rrrr rrrr@{}}
\toprule
& \multicolumn{4}{c}{All documents} & \multicolumn{4}{c}{Two years either side of the cutoff} \\
\cmidrule(lr){2-5}\cmidrule(lr){6-9}
Checkpoint & mean log-prob & Min-K\% & Min-K\%++ & $n$ & mean log-prob & Min-K\% & Min-K\%++ & $n$ \\
\midrule
ChronoGPT, cutoff 2011 & 0.16 & 0.34 & 0.60 & 429 & 0.41 & 0.64 & 0.51 & 65 \\
ChronoGPT, cutoff 2015 & 0.21 & 0.37 & 0.51 & 429 & 0.44 & 0.60 & 0.51 & 64 \\
ChronoGPT, cutoff 2019 & 0.26 & 0.28 & 0.54 & 429 & 0.37 & 0.26 & 0.58 & 66 \\
ChronoGPT, cutoff 2024 & 0.49 & 0.52 & 0.64 & 429 & 0.76 & 0.77 & 0.72 & 50 \\
OLMo-2-7B, against corpus presence & 0.77 & 0.77 & 0.75 & & & & & \\
OLMo-2-13B, against corpus presence & 0.81 & 0.77 & 0.63 & & & & & \\
\bottomrule
\end{tabular}
}
\end{table*}

\end{document}